\documentclass[journal]{IEEEtran}
\ifCLASSOPTIONcompsoc
  \usepackage[nocompress]{cite}
\else
  \usepackage{cite}
\fi
\ifCLASSINFOpdf
    \usepackage[pdftex]{graphicx}
\else
   \usepackage[dvips]{graphicx}
\fi
\usepackage{amsmath}
\usepackage{bm}
\usepackage{array}
\ifCLASSOPTIONcompsoc
  \usepackage[caption=false,font=footnotesize,labelfont=sf,textfont=sf]{subfig}
\else
  \usepackage[caption=false,font=footnotesize]{subfig}
\fi
\usepackage{threeparttable}

\usepackage{fixltx2e}
\usepackage{stfloats}

\usepackage{subfloat}
\usepackage{url}
\usepackage{booktabs}

\usepackage{amsfonts}

\usepackage[colorlinks,linkcolor=blue]{hyperref}

\begin{document}
\title{Tightly-coupled Monocular Visual-odometric SLAM using Wheels and a MEMS Gyroscope}

\author{Meixiang~Quan,
        Songhao~Piao,
        Minglang~Tan,
        Shi-Sheng~Huang
        % <-this % stops a space
\IEEEcompsocitemizethanks{\IEEEcompsocthanksitem S. Piao(piaosh@hit.edu.cn) and M. Quan(15b903042@hit.edu.cn) are the corresponding authors.}}

\maketitle

\begin{abstract}
%\boldmath
In this paper, we present a novel tightly-coupled probabilistic monocular visual-odometric Simultaneous Localization and Mapping algorithm using wheels and a MEMS gyroscope, which can provide accurate, robust and long-term localization for the ground robot moving on a plane. Firstly, we present an odometer preintegration theory that integrates the wheel encoder measurements and gyroscope measurements to a local frame. The preintegration theory properly addresses the manifold structure of the rotation group SO(3) and carefully deals with uncertainty propagation and bias correction. Then the novel odometer error term is formulated using the odometer preintegration model and it is tightly integrated into the visual optimization framework. Furthermore, we introduce a complete tracking framework to provide different strategies for motion tracking when (1) both measurements are available, (2) visual measurements are not available, and (3) wheel encoder experiences slippage, which leads the system to be accurate and robust. Finally, the proposed algorithm is evaluated by performing extensive experiments, the experimental results demonstrate the superiority of the proposed system.
\end{abstract}


% \fi
\IEEEpeerreviewmaketitle

\section{Introduction}
Simultaneous localization and mapping(SLAM) from on-board sensors is a fundamental and key technology for autonomous mobile robot to safely interact within its workspace. SLAM is a technique that builds a globally consistent representation of the environment(i.e. the map) and estimates the state of the robot in the map simultaneously. Because SLAM can be used in many practical applications, such as autonomous driving, virtual or augmented reality and indoor service robots, it has received considerable attention from Robotics and Computer Vision communities.

In this paper, we propose a novel tightly-coupled probabilistic optimization-based monocular visual-odometric SLAM(VOSLAM) system. By combining a monocular camera with wheels and a MEMS gyroscope, the method provides accurate and robust motion tracking for domestic service robots moving on a plane, e.g. cleaning robot, nursing robot and restaurant robot waiter. A single camera provides rich information about the environment, which allows for building 3D map, tracking camera pose and recognizing places already visited. However, the scale of the environment can not be determined using monocular camera, and visual tracking system is sensitive to motion blur, occlusions and illumination changes. Most ground robots are equipped with wheel encoders that provide precise and stable translational measurements of each wheel at most of the time, the measurements contain the absolute scale information. Whereas, the wheel encoder cannot provide accurate self rotational estimates and occasionally provides faulty measurements. In addition, the MEMS gyroscope is a low cost and commercially widely used sensor, and provides accurate and robust inter-frame rotational estimate. However, the estimated rotation is noisy and diverges even in few seconds. Based on the analysis of each sensor, we can know that wheel encoder and gyroscope are complementary to the monocular camera sensor. Therefore, tightly fusing the measurements from wheel encoder and gyroscope to the monocular visual SLAM can not only dramatically improve the accuracy and robustness of the system, but also recover the scale of the environment. In the following, we will call the wheel encoder and MEMS gyroscope the odometer.

In order to tightly fuse the odometer measurements to the visual SLAM system in the framework of graph-based optimization, it is important to provide the integrated odometer measurements between the selected keyframes. Therefore, motivated by the inertial measurement unit(IMU) preintegration theory proposed in \cite{GTSAM}, we present a novel odometer preintegration theory and corresponding uncertainty propagation and bias correction theory on manifold. The preintegration theory integrates the measurements from the wheel encoder and gyroscope to a single relative motion constraint that is independent of the change of the linearization point, therefore the repeated computation is eliminated. Then, based on the proposed odometer preintegration model, we formulate the new preintegrated odometer factor and seamlessly integrate it in a visual-odometric pipeline under the optimization framework.

Furthermore, both visual and odometer measurements are not always available. Therefore, we present a complete visual-odometric tracking framework to ensure the accurate and robust motion tracking in different situations. For the situation where both measurements are available, we maximally exploit the both sensing cues to provide accurate motion tracking. For the situation where visual information is not available, we use odometer measurements to improve the robustness of the system and offer some strategies to render the visual information available as quick as possible. In addition, for the critical drawback of the wheel sensor, we provide a strategy to detect and compensate for the slippage of the wheel encoder. In this way, we can track the motion of the ground robot accurately and robustly.

\begin{figure*}[!t]
\centering
\includegraphics[width=6in]{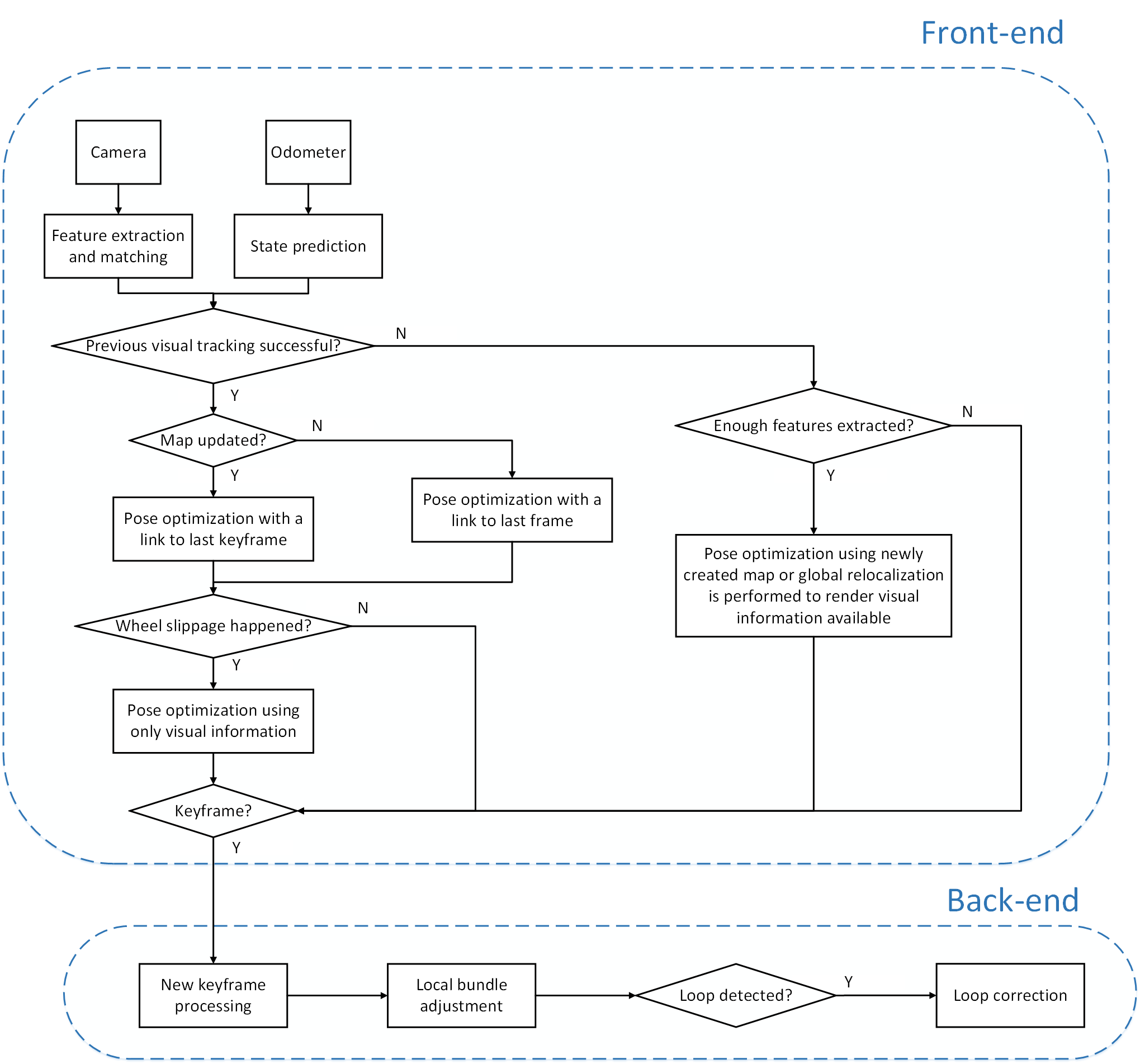}
% where an .eps filename suffix will be assumed under latex,
% and a .pdf suffix will be assumed for pdflatex; or what has been declared
% via \DeclareGraphicsExtensions.
\caption{Flow diagram showing the full pipeline of the proposed system.}
\label{FLOWCHART}
\end{figure*}

The final contribution of the paper is the extensive evaluation of our system. Extensive experiments are performed to demonstrate the accuracy and robustness of the proposed algorithm. The presented algorithm is shown in Fig. \ref{FLOWCHART}, and the details are presented in later sections.

\section{Related work}
There are extensive scholarly works on monocular visual SLAM, these works rely on either filtering methods or nonlinear optimization methods. Filtering based approaches require fewer computational resources due to the continuous marginalization of past state. The first real-time monocular visual SLAM - MonoSLAM \cite{MONOSLAM} is an extended kalman filter(EKF) based method. The standard way of computing Jacobian in the filtering leads the system to have incorrect observability, therefore the system is inconsistent and gets slightly lower accuracy. To solve this problem, the first-estimates Jacobian approach was proposed in \cite{FEJ}, which computes Jacobian with the first-ever available estimate instead of different linearization points to ensure the correct observability of the system and thereby improve the consistency and accuracy of the system. In addition, the observability-constrained EKF \cite{OCEKF} was proposed to explicitly enforce the unobservable directions of the system, hence improving the consistency and accuracy of the system.

On the other hand, nonlinear optimization based approaches can achieve better accuracy due to it's capability to iteratively re-linearize measurement models at each iteration to better deal with their nonlinearity, however it incurs a high computational cost. The first real-time optimization based monocular visual SLAM system is PTAM \cite{ptam} proposed by Klein and Murray. The method achieves real-time performance by dividing the SLAM system into two parallel threads. In one thread, the system performs bundle adjustment over selected keyframes and constructed map points to obtain accurate map of the environment. In the other parallel thread, the camera pose is tracked by minimizing the reprojection error of the features that match the reconstructed map. Based on the work of PTAM, a versatile monocular SLAM system ORB-SLAM \cite{ORBSLAM} was presented. The system introduced the third loop closing thread to eliminate the accumulated error when revisiting an already reconstructed area, it is achieved by taking advantage of bag-of-words \cite{DBOW2} and a 7 degree-of-freedom(dof) pose graph optimization \cite{SCALEDRIFT}.

In addition, according to the definition of visual residual models, monocular SLAM can also be categorized into feature based approaches and direct approaches. The above mentioned methods are all feature based approaches, which is quite mature and able to provide accurate estimate. However, the approaches fail to track in poorly textured environments and need to consume extra computational resources to extract and match features. In contrary, direct methods work on pixel intensity and can exploit all the information in the image even in some places where the gradient is small. Therefore, direct methods can outperform feature based methods in low texture environment and in the case of camera defocus and motion blur. DTAM\cite{DTAM}, SVO\cite{svo} and LSD-SLAM\cite{LSDSLAM} are direct monocular SLAM systems, which builds the dense or semi-dense map from monocular images in real-time, however its accuracy is still lower than the feature based semi-dense mapping technique \cite{ORBDIRECT}.

The monocular visual SLAM is scale ambiguous and sensitive to motion blur, occlusions and illumination changes. Therefore, based on the framework of monocular SLAM, it is often combined with other odometric sensors, especially IMU sensor, to achieve accurate and robust tracking system. Tightly-coupled visual-odometric SLAM can also be categorized into filtering based methods and optimization based methods, where visual and odometric measurements are fused from the raw measurement level. Papers \cite{VISLAMINVDEP,VISLAMGPSDENIED,VINAVIGATION,MSCKF,MSCKF2} are filtering based monocular visual-inertial SLAM, the approaches use the inertial measurements to accurately predict the motion movement between two consecutive frames. An elegant example for filtering based visual-inertial odometry(VIO) is the MSCKF\cite{MSCKF}, which exploits all the available geometric information provided by the visual measurements with the computational complexity only linear in the number of features, it is achieved by excluding point features from the state vector.

OKVIS\cite{OKVIS} is an optimization based monocular visual-inertial SLAM, which tightly integrates the inertial measurements in the keyframe-based visual-inertial pipeline under the framework of graph optimization. However, in this system, the IMU integration is computed repeatedly when the linearization point changes. Therefore in order to eliminate this repeated computation, Forster et al. presented an IMU preintegration theory, and tightly integrated the preintegrated IMU factor and visual factor in a fully probabilistic manner in paper \cite{GTSAM}. Later, a real-time tightly-coupled monocular visual-inertial SLAM system - ORB-VISLAM \cite{ORBVIO} was presented. The system can close loop and reuse the previously estimated 3D map, therefore achieve high accuracy and robustness. Recently, another tightly-coupled monocular visual-inertial odometry was proposed in \cite{VINSMONO1}\cite{VINSMONO2}, it provides accurate and robust motion tracking by performing local bundle adjustment(BA) for each frame and its capability to close loop.

There are also several works on the visual-odometric SLAM that fuses visual measurements and wheel encoder measurements. In \cite{1point}, wheel encoder measurements are combined to the system of visual motion tracking for accurate motion prediction, thereby the true scale of the system is recovered. In addition, paper \cite{VINSWHEEL} proved that VINS has additional unobservable directions when a ground robot moves along straight lines or circular arcs. Therefore a system fusing the wheel encoder measurements to the VINS estimator in a tightly-coupled manner was proposed to render the scale of the system observable.

\section{Preliminaries}\label{preliminaries}

We begin by briefly defining the notations used throughout the paper. We employ $(\cdot)^W$ to denote the world reference frame, $(\cdot)^{O_k}$, $(\cdot)^{C_k}$ and $(\cdot)^{B_k}$ to denote the wheel odometer frame, camera frame and inertial frame for the $k^{th}$ image. In following, we employ $\bm{\mathrm{R}}^{\mathcal{F}_1}_{\mathcal{F}_2} \in \bm{\mathrm{SO}}(3)$ to represent rotation from frame $\{\mathcal{F}_2\}$ to $\{\mathcal{F}_1\}$ and ${\bm{\mathrm{p}}}^{\mathcal{F}_1}_{\mathcal{F}_2} \in \mathbb{R}^3$ to describe the 3D position of frame $\{\mathcal{F}_2\}$ with respect to the frame $\{\mathcal{F}_1\}$.

The rotation and translation between the rigidly mounted wheel encoder and camera sensor are ${\bm{\mathrm{R}}^C_O \in \bm{\mathrm{SO}}(3)}$ and $\bm{\mathrm{p}}^C_O \in \mathbb{R}^3$ respectively, and ${\bm{\mathrm{R}}_B^O \in \bm{\mathrm{SO}}(3)}$ denotes the rotation from the inertial frame to the wheel encoder frame, these parameters are obtained from calibration. In addition, the pose of the $k^{th}$ image is the rigid-body transformation $\bm{\mathrm{T}}^{O_k}_W = \left[ \begin{array}{cc}  \bm{\mathrm{R}}^{O_k}_W & \bm{\mathrm{p}}^{O_k}_W \\  \bm{0}^T & 1 \end{array} \right] \in \bm{\mathrm{SE}}(3)$, and the 3D position of the $j^{th}$ map point in the global frame $\{W\}$ and the camera frame $\{C_k\}$ are denoted as $\bm{\mathrm{f}}^W_j \in \mathbb{R}^3$ and $\bm{\mathrm{f}}^{C_k}_j \in \mathbb{R}^3$ respectively.

In order to provide a minimal representation for the rigid-body transformation during the optimization, we use a vector $\bm{\xi} \in \mathbb{R}^3$ computed from the Lie algebra of $\bm{\mathrm{SO}}(3)$ to represent the over-parameterized rotation matrix $\bm{\mathrm{R}}$. The Lie algebra of $\bm{\mathrm{SO}}(3)$ is denoted as $\mathfrak{so}(3)$, which is the tangent space of the manifold and coincides with the space of $3 \times 3$ skew symmetric matrices. The logarithm map associates a rotation matrix $\bm{\mathrm{R}} \in \bm{\mathrm{SO}}(3)$ to a skew symmetric matrix:

\begin{equation}
    \bm{\xi}^{\wedge} = \mathrm{log}(\bm{\mathrm{R}})
\end{equation}
where ${(\cdot)}^{\wedge}$ operator maps a 3-dimensional vector to a skew symmetric matrix, thus the vector $\bm{\xi}$ can be computed using inverse ${(\cdot)}^{\vee}$ operator:

\begin{equation}
    \bm{\xi} = \mathrm{Log}(\bm{\mathrm{R}}) = {\mathrm{log}(\bm{\mathrm{R}})}^{\vee}
\end{equation}

Inversely, the exponential map associates the Lie algebra $\mathfrak{so}(3)$ to the rotation matrix $\bm{\mathrm{R}} \in \bm{\mathrm{SO}}(3)$:
\begin{equation}
    \bm{\mathrm{R}} = \mathrm{Exp}(\bm{\xi}) = \mathrm{exp}({\bm{\xi}}^{\wedge})
\end{equation}

The input of our estimation problem is a stream of measurements from the monocular camera and the odometer. The visual measurement is a set of point features extracted from the captured intensity image $\mathrm{I}_k: \Omega \subset \mathbb{R}^2 \to \mathbb{R}$ at time-step $k$. Such measurement is obtained by camera projection model $\pi: \mathbb{R}^3 \to \mathbb{R}^2$, which projects the $l^{th}$ map point expressed in the current camera frame $\bm{\mathrm{f}}^{C_k}_l=(x_c, y_c, z_c)^\mathrm{T} \in \mathbb{R}^3$ onto the image coordinate $\bm{\mathrm{z}}_{kl} = (u,v)^\mathrm{T} \in \Omega$:
\begin{equation}
\begin{split}
\widetilde{\bm{\mathrm{z}}}_{kl} &= \bm{\mathrm{z}}_{kl} + \bm{\sigma}_{kl}\\
&=\pi(\bm{\mathrm{f}}^{C_k}_l) + \bm{\sigma}_{kl}\\
\label{visualmm}
\end{split}
\end{equation}
where $\widetilde{\bm{\mathrm{z}}}_{kl}$ is the corresponding feature measurement, and $\bm{\sigma}_{kl}$ is the $2 \times 1$ measurement noise with covariance $\bm{\Sigma}_{C_{kl}}$. The projection function $\pi$ is determined by the intrinsic parameters of the camera, which is known from calibration.

In addition, the gyroscope of the odometer measures the angular velocity $\widetilde{\bm{\mathrm{\omega}}}_{k}$ at time-step k, the measurement is assumed to be affected by a slowly varying sensor bias $\bm{\mathrm{b}}_{g_k}$ with covariance $\bm{\Sigma}_{b_g}$ and a discrete-time zero-mean Gaussian white noise $\bm{\mathrm{\eta}}_{gd}$ with covariance $\bm{\Sigma}_{gd}$:
\begin{equation}
    \widetilde{\bm{\mathrm{\omega}}}_{k} = \bm{\mathrm{\omega}}_k + \bm{\mathrm{b}}_{g_k} + \bm{\mathrm{\eta}}_{gd}
\end{equation}

The wheel encoder of the odometer measures the traveled distance $\widetilde{\mathrm{Dl}}_k$ and $\widetilde{\mathrm{Dr}}_{k}$ of the both wheels from time-step $k-1$ to $k$, which is assumed to be affected by a discrete-time zero-mean Gaussian white noise $\mathrm{\eta}_{ed}$ with variance ${\Sigma}_{ed}$:
\begin{equation}
\begin{split}
\widetilde{\mathrm{Dl}}_{k} &= \mathrm{Dl}_k + \mathrm{\eta}_{ed}\\
\widetilde{\mathrm{Dr}}_{k}&= \mathrm{Dr}_k + \mathrm{\eta}_{ed}\\
\end{split}
\end{equation}

Therefore, the measured 3D position of frame $\{O_{k}\}$ with respect to frame $\{O_{k-1}\}$ from wheel encoder is:
\begin{equation}
\begin{split}
\widetilde{\bm{\mathrm{\psi}}}^{O_{k-1}}_{O_k} &= \bm{\mathrm{\psi}}^{O_{k-1}}_{O_k} + \bm{\mathrm{\eta}}_{\psi d}\\
\left[ \begin{array}{c} \cfrac{\widetilde{\mathrm{Dl}}_{k}+\widetilde{\mathrm{Dr}}_{k}}{2} \\ 0 \\ 0 \end{array} \right] &=  -\bm{\mathrm{R}}^{O_{k-1}}_W {\bm{\mathrm{R}}^{O_k}_W}^\mathrm{T} \bm{\mathrm{p}}^{O_k}_W + \bm{\mathrm{p}}^{O_{k-1}}_W + \bm{\mathrm{\eta}}_{\psi d}(\mathrm{\eta}_{ed})
\end{split}
\end{equation}
where $\bm{\mathrm{R}}^{O_{k-1}}_W$ and $\bm{\mathrm{p}}^{O_{k-1}}_W$ constitute the pose of frame $\{O_{k-1}\}$, and $\bm{\mathrm{R}}^{O_{k}}_W$ and $\bm{\mathrm{p}}^{O_{k}}_W$ constitute the pose of frame $\{O_{k}\}$.

In many cases, the ground robot is moving on a plane. The motion on a plane has 3 dof in contrast to 6 dof of 3D motion, i.e. the roll, pitch angle and translation on z-axis of frame $\{O_k\}$ in the frame of physical plane should be close to zero. Since the additional information can improve the accuracy of the system, we also provide planar measurement $\widetilde{\bm{\mathrm{pl}}}_k = [0, 0, 0]^\mathrm{T} \in \mathbb{R}^3$ with covariance $\bm{\Sigma}_\mathrm{pl}$ for each frame, where the first two elements correspond to the planar rotational measurement and the third element corresponds to the planar translational measurement.

\section{Tightly-coupled visual-odometric nonlinear optimization on Manifold}

We use $\mathcal{K}$ to denote the set of successive keyframes from $\mathfrak{i}$ to $\mathfrak{j}$, and $\mathcal{L}$ to denote all the landmarks visible from the keyframes in $\mathcal{K}$. Then the variables to be estimated in the window of keyframes from $\mathfrak{i}$ to $\mathfrak{j}$ is:
\begin{equation}
    \bm{\mathcal{X}} = \{\bm{x}_k, \bm{\mathrm{f}}^W_l \}_{k \in \mathcal{K}, l \in \mathcal{L}}
\end{equation}
where $\bm{x}_k = \{\bm{\mathrm{T}}^{O_k}_W, \bm{\mathrm{b}}_{g_k} \}$ is the state of the keyframe $k$.

We denote the visual measurements of $\mathcal{L}$ at the keyframe $i$ as $\bm{\mathcal{Z}}_{C_i} = \{\widetilde{\bm{\mathrm{z}}}_{il}\}_{l \in \mathcal{L}}$. In addition, we denote the odometer measurements obtained between two consecutive keyframes $i$ and $j$ as $\bm{\mathcal{O}}_{ij} = \{ \widetilde{\bm{\mathrm{\omega}}}_{t},  \widetilde{\mathrm{Dl}}_{t}, \widetilde{\mathrm{Dr}}_{t} \}_{\mathrm{time-step}(i) \leq t < \mathrm{time-step}(j)}$. Therefore, the set of measurements collected for optimizing the state $\bm{\mathcal{X}}$ is:
\begin{equation}
    \bm{\mathcal{Z}} = \{\bm{\mathcal{Z}}_{C_i}, \bm{\mathcal{O}}_{ij}, \widetilde{\bm{\mathrm{pl}}}_i\}_{(i,j) \in \mathcal{K}}
\end{equation}

\subsection{Maximum a Posteriori Estimation}
The optimum value of state $\bm{\mathcal{X}}$ is estimated by solving the following maximum a posteriori (MAP) problem:
\begin{equation}
    \bm{\mathcal{X}} ^\ast = \mathop{\mathrm{argmax}} \limits_{\bm{\mathcal{X}}}\ p(\bm{\mathcal{X}}|\bm{\mathcal{Z}})
\label{MAP}
\end{equation}
which means that given the available measurements $\bm{\mathcal{Z}}$, we want to find the best estimate for state $\bm{\mathcal{X}}$. Assuming measurements $\bm{\mathcal{Z}}$ are independent, then using Bayes' rule, we can rewrite $p(\bm{\mathcal{X}}|\bm{\mathcal{Z}})$ as:

\begin{equation}
\begin{split}
&p(\bm{\mathcal{X}}|\bm{\mathcal{Z}}) \propto p(\bm{\mathcal{X}}_0)p(\bm{\mathcal{Z}}|\bm{\mathcal{X}})\\
&= p(\bm{\mathcal{X}}_0)\prod \limits_{(i,j) \in \mathcal{K}} p(\bm{\mathcal{Z}}_{C_i}, \bm{\mathcal{O}}_{ij}, \widetilde{\bm{\mathrm{pl}}}_i|\bm{\mathcal{X}})\\
&= p(\bm{\mathcal{X}}_0)\prod \limits_{(i,j) \in \mathcal{K}} p(\bm{\mathcal{O}}_{ij}|\bm{x}_i,\bm{x}_j)\prod \limits_{i \in \mathcal{K}}\prod \limits_{l \in \mathcal{L}} p(\widetilde{\bm{\mathrm{z}}}_{il}|\bm{x}_i, \bm{\mathrm{f}}^W_l) \prod \limits_{i \in \mathcal{K}} p(\widetilde{\bm{\mathrm{pl}}}_i|\bm{x}_i)
\end{split}
\end{equation}

\begin{figure}[!t]
\centering
\includegraphics[width=3.5in]{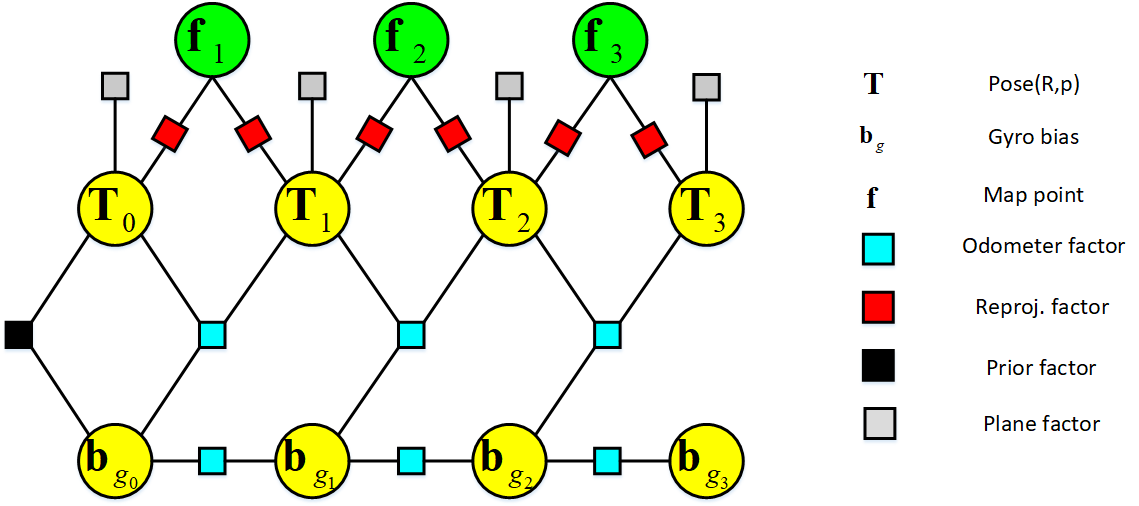}
\caption{Factor graph representing the visual-odometric optimization problem. The states are shown as circles and factors are shown as squares. The blue squares represent the odometer factors and connect to the state of the previous keyframe, red squares denote the visual factors corresponding to camera observations, black squares denote prior factors and gray squares denote the plane factors.}
\label{factorgraph}
\end{figure}

The equation can be interpreted as a factor graph. The variables in $\bm{\mathcal{X}}$ are corresponding to nodes in the factor graph. The terms $p(\bm{\mathcal{X}}_0)$, $p(\bm{\mathcal{O}}_{ij}|\bm{x}_i,\bm{x}_j)$, $p(\widetilde{\bm{\mathrm{z}}}_{il}|\bm{x}_i, \bm{\mathrm{f}}^W_l)$ and $p(\widetilde{\bm{\mathrm{pl}}}_i|\bm{x}_i)$ are called factors, which encodes probabilistic constraints between nodes. A factor graph representing the problem is shown in Fig. \ref{factorgraph}.

The MAP estimate is equal to the minimum of the negative log-posterior. Under the assumption of zero-mean Gaussian noise, the MAP estimate in \eqref{MAP} can be written as the minimization of sum of the squared residual errors:
\begin{equation}
\begin{split}
\bm{\mathcal{X}} ^\ast = &\mathop{\mathrm{argmin}} \limits_{\bm{\mathcal{X}}}\ - \mathrm{log}\ p(\bm{\mathcal{X}}|\bm{\mathcal{Z}}) \\
= &\mathop{\mathrm{argmin}} \limits_{\bm{\mathcal{X}}} \|\bm{\mathrm{r}}_0\|^2_{{\bm{\Sigma}}_0}
+ \sum \limits_{(i,j) \in \mathcal{K}} \rho\left(\|{\bm{\mathrm{r}}_{\mathcal{O}_{ij}}}\|^2_{{\bm{\Sigma}}_{\mathcal{O}_{ij}}}\right)\\
&+ \sum \limits_{i \in \mathcal{K}}\sum \limits_{l \in \mathcal{L}} \rho \left( \| {\bm{\mathrm{r}}_{\mathcal{C}_{il}}}\|^2_{{\bm{\Sigma}}_{\mathcal{C}_{il}}} \right) + \sum \limits_{i \in \mathcal{K}} \rho\left(\|{\bm{\mathrm{r}}_{\mathrm{pl}_i}}\|^2_{{\bm{\Sigma}}_{\mathrm{pl}}}\right)
\label{MAPfactor}
\end{split}
\end{equation}
where $\bm{\mathrm{r}}_0$, $\bm{\mathrm{r}}_{\mathcal{O}_{ij}}$, $\bm{\mathrm{r}}_{\mathcal{C}_{il}}$ and $\bm{\mathrm{r}}_{\mathrm{pl}_i}$ are the prior error, odometer error, reprojection error and plane error respectively, as well as $\bm{\Sigma}_0$, ${\bm{\Sigma}}_{\mathcal{O}_{ij}}$, ${\bm{\Sigma}}_{\mathcal{C}_{il}}$ and ${\bm{\Sigma}}_{\mathrm{pl}}$ are the corresponding covariance matrices, and $\rho$ is the Huber robust cost function. In the following subsections, we provide expressions for these residual errors and introduce the Gauss-Newton optimization method on manifold.

\subsection{Preintegrated Odometer Measurement}
In this section, we derive the odometer preintegration between two consecutive keyframes $i$ and $j$ by assuming the gyro bias of keyframe $i$ is known. We firstly define the rotation increment $\Delta \bm{\mathrm{R}}_{ij}$ and position increment $\Delta \bm{\mathrm{p}}_{ij}$ in the wheel odometer frame $\{O_i\}$ as:
\begin{equation}
\begin{split}
\Delta \bm{\mathrm{R}}_{ij} &= \prod_{k=i}^{j-1} \bm{\mathrm{R}}_B^O \mathrm{Exp}\left((\widetilde{\bm{\mathrm{\omega}}}_{k} - \bm{\mathrm{b}}_{g_i} - \bm{\mathrm{\eta}}_{gd}) \Delta t \right) {\bm{\mathrm{R}}_B^O}^\mathrm{T} \\
\Delta \bm{\mathrm{p}}_{ij} &= \sum_{k=i+1}^{j} \Delta \bm{\mathrm{R}}_{ik-1}(\widetilde{\bm{\mathrm{\psi}}}_{O_k}^{O_{k-1}} - \bm{\mathrm{\eta}}_{\psi d})\\
\label{increment}
\end{split}
\end{equation}

Then, using the first-order approximation and dropping higher-order noise terms, we split each increment in \eqref{increment} to preintegrated measurement and its noise. For rotation, we have:
\begin{equation}
\begin{split}
&\Delta \bm{\mathrm{R}}_{ij} = \prod_{k=i}^{j-1} \mathrm{Exp}\left( \bm{\mathrm{R}}_B^O (\widetilde{\bm{\mathrm{\omega}}}_{k} - \bm{\mathrm{b}}_{g_i}  - \bm{\mathrm{\eta}}_{gd}) \Delta t \right)\\
&= \prod_{k=i}^{j-1} \left[ \mathrm{Exp}\left( \bm{\mathrm{R}}_B^O (\widetilde{\bm{\mathrm{\omega}}}_{k} - \bm{\mathrm{b}}_{g_i}) \Delta t \right) \mathrm{Exp}\left( - \bm{\mathrm{J}}_{r_k} \bm{\mathrm{R}}_B^O \bm{\mathrm{\eta}}_{gd} \Delta t \right) \right] \\
&= \Delta \widetilde{\bm{\mathrm{R}}}_{ij} \prod_{k=i}^{j-1} \mathrm{Exp}\left( -\Delta \widetilde{\bm{\mathrm{R}}}_{k+1j}^\mathrm{T} \bm{\mathrm{J}}_{r_k} \bm{\mathrm{R}}_B^O \bm{\mathrm{\eta}}_{gd} \Delta t \right)\\
&= \Delta \widetilde{\bm{\mathrm{R}}}_{ij} \mathrm{Exp}(-\delta \bm{\mathrm{\phi}}_{ij})
\label{rotprenoise}
\end{split}
\end{equation}
where $\bm{\mathrm{J}}_{r_k} = \bm{\mathrm{J}}_r(\bm{\mathrm{R}}_B^O (\widetilde{\bm{\mathrm{\omega}}}_{k} - \bm{\mathrm{b}}_{g_i}) \Delta t)$. Therefore, we obtain the preintegrated rotation measurement:
\begin{equation}
\Delta \widetilde{\bm{\mathrm{R}}}_{ij} = \prod_{k=i}^{j-1} \mathrm{Exp}\left( \bm{\mathrm{R}}_B^O (\widetilde{\bm{\mathrm{\omega}}}_{k} - \bm{\mathrm{b}}_{g_i}) \Delta t \right)
\end{equation}

For position, we have:
\begin{equation}
\begin{split}
&\Delta \bm{\mathrm{p}}_{ij} = \sum_{k=i+1}^{j} \left[ \Delta\widetilde{\bm{\mathrm{R}}}_{ik-1}(\bm{\mathrm{I}}-\delta \bm{\mathrm{\phi}}_{ik-1}^{\wedge})\widetilde{\bm{\mathrm{\psi}}}_{O_k}^{O_{k-1}} - \Delta\widetilde{\bm{\mathrm{R}}}_{ik-1} \bm{\mathrm{\eta}}_{\psi d} \right]\\
&= \Delta \widetilde{\bm{\mathrm{p}}}_{ij} + \sum_{k=i+1}^{j} \left[ \Delta\widetilde{\bm{\mathrm{R}}}_{ik-1}
{\widetilde{\bm{\mathrm{\psi}}}^{{O_{k-1}}^{\wedge}}_{O_k}} \delta \bm{\mathrm{\phi}}_{ik-1} - \Delta\widetilde{\bm{\mathrm{R}}}_{ik-1} \bm{\mathrm{\eta}}_{\psi d} \right] \\
&= \Delta \widetilde{\bm{\mathrm{p}}}_{ij} - \delta \bm{\mathrm{p}}_{ij}
\label{posprenoise}
\end{split}
\end{equation}

Therefore, we obtain the preintegrated position measurement:
\begin{equation}
\Delta \widetilde{\bm{\mathrm{p}}}_{ij} = \sum_{k=i+1}^{j} \Delta\widetilde{\bm{\mathrm{R}}}_{ik-1} \widetilde{\bm{\mathrm{\psi}}}_{O_k}^{O_{k-1}}
\end{equation}

\subsection{Noise Propagation}
We start with rotation noise. From \eqref{rotprenoise}, we obtain:
\begin{equation}
\delta \bm{\mathrm{\phi}}_{ij} = \sum_{k=i}^{j-1} \Delta \widetilde{\bm{\mathrm{R}}}_{k+1j}^\mathrm{T} \bm{\mathrm{J}}_{r_k} \bm{\mathrm{R}}_B^O \bm{\mathrm{\eta}}_{gd} \Delta t
\label{rotnoise}
\end{equation}

The rotation noise term $\delta \bm{\mathrm{\phi}}_{ij}$ is zero-mean and Gaussian, since it is a linear combination of zero-mean white Gaussian noise $\bm{\eta}_{gd}$.

Furthermore, from \eqref{posprenoise}, we obtain the position noise:
\begin{equation}
\delta \bm{\mathrm{p}}_{ij} = \sum_{k=i+1}^{j} \left[- \Delta \widetilde{\bm{\mathrm{R}}}_{ik-1} {\widetilde{\bm{\mathrm{\psi}}}^{{O_{k-1}}^{\wedge}}_{O_k}} \delta \bm{\mathrm{\phi}}_{ik-1}+ \Delta\widetilde{\bm{\mathrm{R}}}_{ik-1} \bm{\mathrm{\eta}}_{\psi d} \right]
\label{posnoise}
\end{equation}

The position noise $\delta \bm{\mathrm{p}}_{ij} $ is also zero-mean Gaussian noise, because it is a linear combination of the noise $\bm{\mathrm{\eta}}_{\psi d}$ and the rotation noise $\delta \bm{\mathrm{\phi}}_{ij}$.

We write \eqref{rotnoise} and \eqref{posnoise} in iterative form, then the noise propagation can be written in matrix form as:
\begin{equation}
\begin{split}
\left[ \begin{array}{c} \delta \bm{\mathrm{\phi}}_{ik+1} \\ \delta \bm{\mathrm{p}}_{ik+1} \end{array} \right] &= \left[ \begin{array}{cc} \Delta \widetilde{\bm{\mathrm{R}}}_{kk+1}^\mathrm{T} & \bm{\mathrm{0}}_{3 \times 3} \\ -\Delta \widetilde{\bm{\mathrm{R}}}_{ik} {\widetilde{\bm{\mathrm{\psi}}}^{{O_{k-1}}^{\wedge}}_{O_k}}& \bm{\mathrm{I}}_{3 \times 3} \end{array} \right] \left[ \begin{array}{c} \delta \bm{\mathrm{\phi}}_{ik} \\ \delta \bm{\mathrm{p}}_{ik} \end{array} \right] \\
&+ \left[ \begin{array}{cc} \bm{\mathrm{J}}_{r_k} \bm{\mathrm{R}}_B^O  \Delta t & \bm{\mathrm{0}}_{3 \times 3} \\ \bm{\mathrm{0}}_{3 \times 3} & \Delta \widetilde{\bm{\mathrm{R}}}_{ik} \end{array} \right] \left[ \begin{array}{c} \bm{\mathrm{\eta}}_{gd} \\ \bm{\mathrm{\eta}}_{\psi d} \end{array} \right]
\end{split}
\end{equation}
or more simply:
\begin{equation}
\bm{\mathrm{\eta}}_{ik+1} = \bm{\mathrm{A}}\bm{\mathrm{\eta}}_{ik} + \bm{\mathrm{B}}\bm{\mathrm{\eta}}_{d}
\label{linearmodel}
\end{equation}

Given the linear model \eqref{linearmodel} and the covariance $\bm{\Sigma}_{\eta_d} \in \mathbb{R}^{6 \times 6}$ of the odometer measurements noise $\bm{\mathrm{\eta}}_{d}$, it is possible to compute the covariance of the odometer preintegration noise iteratively:
\begin{equation}
\bm{\mathrm{\Sigma}}_{ik+1} = \bm{\mathrm{A}}\bm{\mathrm{\Sigma}}_{ik}\bm{\mathrm{A}}^\mathrm{T} + \bm{\mathrm{B}}\bm{\Sigma}_{\eta_d}\bm{\mathrm{B}}^\mathrm{T}
\end{equation}
with initial condition $\bm{\mathrm{\Sigma}}_{ii} = \bm{\mathrm{0}}_{6 \times 6}$.

Therefore, we can fully characterize the preintegrated odometer measurements noise as:
\begin{equation}
\bm{\mathrm{\eta}}_{ij} = {[ \begin{array}{cc} \delta \bm{\mathrm{\phi}}_{ij}^\mathrm{T} & \delta \bm{\mathrm{p}}_{ij}^\mathrm{T} \end{array} ]}^\mathrm{T} \sim \mathcal{N}(\bm{\mathrm{0}}_{6 \times 1}, \bm{\mathrm{\Sigma}}_{ij})
\end{equation}

\subsection{Bias update}
In the previous section, we assumed that the gyro bias $\bm{\mathrm{b}}_{g_i}$ is fixed. Given the bias change $\delta \bm{\mathrm{b}}_{g}$, we can update the preintegrated measurements using a first-order approximation. For preintegrated rotation measurement:
\begin{equation}
\begin{split}
&\Delta \widetilde{\bm{\mathrm{R}}}_{ij}(\bm{\mathrm{b}}_{g_i}) = \prod_{k=i}^{j-1} \mathrm{Exp}\left( \bm{\mathrm{R}}_B^O (\widetilde{\bm{\mathrm{\omega}}}_{k} - \Bar{\bm{\mathrm{b}}}_{g_i}  - \delta \bm{\mathrm{b}}_{g}) \Delta t \right)\\
&= \prod_{k=i}^{j-1} \left[ \mathrm{Exp}\left( \bm{\mathrm{R}}_B^O (\widetilde{\bm{\mathrm{\omega}}}_{k} - \bm{\mathrm{b}}_{g_i}) \Delta t \right) \mathrm{Exp}\left( - \bm{\mathrm{J}}_{r_k} \bm{\mathrm{R}}_B^O \delta \bm{\mathrm{b}}_{g} \Delta t \right) \right]\\
&= \Delta \widetilde{\bm{\mathrm{R}}}_{ij}(\Bar{\bm{\mathrm{b}}}_{g_i}) \prod_{k=i}^{j-1} \mathrm{Exp}\left( - \Delta \widetilde{\bm{\mathrm{R}}}_{k+1j}^\mathrm{T} \bm{\mathrm{J}}_{r_k} \bm{\mathrm{R}}_B^O \delta \bm{\mathrm{b}}_{g} \Delta t \right)\\
&= \Delta \widetilde{\bm{\mathrm{R}}}_{ij}(\Bar{\bm{\mathrm{b}}}_{g_i}) \mathrm{Exp}(\cfrac{\partial \Delta \bar{\bm{\mathrm{R}}}_{ij}}{\partial \bm{\mathrm{b}}_{g}} \delta \bm{\mathrm{b}}_{g})
\end{split}
\end{equation}
where $\cfrac{\partial \Delta \bar{\bm{\mathrm{R}}}_{ij}}{\partial \bm{\mathrm{b}}_{g}} = \sum_{k=i}^{j-1} - \Delta \widetilde{\bm{\mathrm{R}}}_{k+1j}^\mathrm{T} \bm{\mathrm{J}}_{r_k} \bm{\mathrm{R}}_B^O  \Delta t$. For preintegrated position measurement:
\begin{equation}
\begin{split}
&\Delta \widetilde{\bm{\mathrm{p}}}_{ij}(\bm{\mathrm{b}}_{g_i}) = \sum_{k=i+1}^{j} \Delta \widetilde{\bm{\mathrm{R}}}_{ik-1}(\Bar{\bm{\mathrm{b}}}_{g_i}) \mathrm{Exp} \left(\cfrac{\partial \Delta \bar{\bm{\mathrm{R}}}_{ik-1}}{\partial \bm{\mathrm{b}}_{g}} \delta \bm{\mathrm{b}}_{g} \right) \widetilde{\bm{\mathrm{\psi}}}_{O_k}^{O_{k-1}}\\
&= \sum_{k=i+1}^{j} \Delta \widetilde{\bm{\mathrm{R}}}_{ik-1}(\Bar{\bm{\mathrm{b}}}_{g_i}) \left(\bm{\mathrm{I}} + {\left(\cfrac{\partial \Delta \bar{\bm{\mathrm{R}}}_{ik-1}}{\partial \bm{\mathrm{b}}_{g}} \delta \bm{\mathrm{b}}_{g}\right)}^{\wedge} \right) \widetilde{\bm{\mathrm{\psi}}}_{O_k}^{O_{k-1}}\\
&= \Delta \widetilde{\bm{\mathrm{p}}}_{ij}(\Bar{\bm{\mathrm{b}}}_{g_i}) - \sum_{k=i+1}^{j} \Delta \widetilde{\bm{\mathrm{R}}}_{ik-1}(\Bar{\bm{\mathrm{b}}}_{g_i}) {\widetilde{\bm{\mathrm{\psi}}}^{{O_{k-1}}^{\wedge}}_{O_k}} \cfrac{\partial \Delta \bar{\bm{\mathrm{R}}}_{ik-1}}{\partial \bm{\mathrm{b}}_{g}} \delta \bm{\mathrm{b}}_{g}\\
&= \Delta \widetilde{\bm{\mathrm{p}}}_{ij}(\Bar{\bm{\mathrm{b}}}_{g_i}) + \cfrac{\partial \Delta \bar{\bm{\mathrm{p}}}_{ij}}{\partial \bm{\mathrm{b}}_{g}} \delta \bm{\mathrm{b}}_{g}\\
\end{split}
\end{equation}
where $\cfrac{\partial \Delta \bar{\bm{\mathrm{p}}}_{ij}}{\partial \bm{\mathrm{b}}_{g}} = -\sum_{k=i+1}^{j} \Delta \widetilde{\bm{\mathrm{R}}}_{ik-1}(\Bar{\bm{\mathrm{b}}}_{g_i}) {\widetilde{\bm{\mathrm{\psi}}}^{{O_{k-1}}^{\wedge}}_{O_k}} \cfrac{\partial \Delta \bar{\bm{\mathrm{R}}}_{ik-1}}{\partial \bm{\mathrm{b}}_{g}}$.

\subsection{Preintegrated Odometer Measurement Model}

From the geometric relations between two consecutive keyframes $i$ and $j$, we get our preintegrated measurement model as:
\begin{equation}
\begin{split}
\Delta \widetilde{\bm{\mathrm{R}}}_{ij}(\bm{\mathrm{b}}_{g_i}) &= \bm{\mathrm{R}}^{O_i}_W {\bm{\mathrm{R}}^{O_j}_W}^\mathrm{T} \mathrm{Exp}(\delta \bm{\mathrm{\phi}}_{ij})\\
\Delta \widetilde{\bm{\mathrm{p}}}_{ij}(\bm{\mathrm{b}}_{g_i}) &= -\bm{\mathrm{R}}^{O_i}_W {\bm{\mathrm{R}}^{O_j}_W}^\mathrm{T} \bm{\mathrm{p}}^{O_j}_W + \bm{\mathrm{p}}^{O_i}_W + \delta \bm{\mathrm{p}}_{ij}\\
\label{mesuremodel}
\end{split}
\end{equation}

Therefore, preintegrated odometer residual $\bm{\mathrm{r}}_{\Delta_{ij}} =
{\left[ \bm{\mathrm{r}}_{\Delta \bm{\mathrm{R}}_{ij}}^\mathrm{T}, \bm{\mathrm{r}}_{\Delta \bm{\mathrm{p}}_{ij}}^\mathrm{T} \right]}^\mathrm{T} \in \mathbb{R}^6$ is:
\begin{equation}
\begin{split}
&\bm{\mathrm{r}}_{\Delta \bm{\mathrm{R}}_{ij}} = \mathrm{Log} \left( \Delta \widetilde{\bm{\mathrm{R}}}_{ij}(\Bar{\bm{\mathrm{b}}}_{g_i}) \mathrm{Exp}(\cfrac{\partial \Delta \bar{\bm{\mathrm{R}}}_{ij}}{\partial \bm{\mathrm{b}}_{g}} \delta \bm{\mathrm{b}}_{g}) \bm{\mathrm{R}}^{O_j}_W {\bm{\mathrm{R}}^{O_i}_W}^\mathrm{T} \right) \\
&\bm{\mathrm{r}}_{\Delta \bm{\mathrm{p}}_{ij}} = -\bm{\mathrm{R}}^{O_i}_W {\bm{\mathrm{R}}^{O_j}_W}^\mathrm{T} \bm{\mathrm{p}}^{O_j}_W + \bm{\mathrm{p}}^{O_i}_W - (\Delta \widetilde{\bm{\mathrm{p}}}_{ij}(\Bar{\bm{\mathrm{b}}}_{g_i}) + \cfrac{\partial \Delta \bar{\bm{\mathrm{p}}}_{ij}}{\partial \bm{\mathrm{b}}_{g}} \delta \bm{\mathrm{b}}_{g})
\label{odofactor}
\end{split}
\end{equation}

%The derivation of the Jacobian can be seen in appendix.

\subsection{Gyro Bias Model}
Gyro bias is slowly time-varying, so the relation of gyro bias between two consecutive keyframes $i$ and $j$ is:
\begin{equation}
    \bm{\mathrm{b}}_{g_j} = \bm{\mathrm{b}}_{g_i} + \bm{\mathrm{\eta}}_{b_{gd}}
\end{equation}
where $\bm{\mathrm{\eta}}_{b_{gd}}$ is the discrete-time zero-mean Gaussian noise with covariance $\bm{\Sigma}_{b_{gd}}$. Therefore, we can express the gyro bias residual $\bm{\mathrm{r}}_{b_g} \in \mathbb{R}^3$ as:
\begin{equation}
   \bm{\mathrm{r}}_{b_g} =  \bm{\mathrm{b}}_{g_j} -\bm{\mathrm{b}}_{g_i}
 \label{biasfactor}
\end{equation}

\subsection{Odometer Factor}
Given the preintegrated odometer residual in \eqref{odofactor} and the gyro bias residual in \eqref{biasfactor}, the odometer error term in \eqref{MAPfactor} is:
\begin{equation}
    \|{\bm{\mathrm{r}}_{\mathcal{O}_{ij}}}\|^2_{{\bm{\Sigma}}_{\mathcal{O}_{ij}}} = \bm{\mathrm{r}}_{\Delta_{ij}}^\mathrm{T} * \bm{\mathrm{\Sigma}}_{ij}^{-1} * \bm{\mathrm{r}}_{\Delta_{ij}} + \bm{\mathrm{r}}_{b_g}^\mathrm{T} * \bm{\mathrm{\Sigma}}_{b_{gd}}^{-1} * \bm{\mathrm{r}}_{b_g}
\end{equation}

\subsection{Visual Factor}
Through the measurement model in \eqref{visualmm}, the $l^{th}$ map point expressed in the world reference frame $\{W\}$ can be projected onto the image plane of the $i^{th}$ keyframe as:
\begin{equation}
    \bm{\mathrm{z}}_{il}=\pi(\bm{\mathrm{R}}^{C}_O \bm{\mathrm{R}}^{O_i}_W * \bm{\mathrm{f}}^W_l + \bm{\mathrm{R}}^{C}_O  \bm{\mathrm{p}}^{O_i}_W + \bm{\mathrm{p}}^{C}_O  )
\end{equation}
Therefore, the reprojection error $\bm{\mathrm{r}}_{\mathcal{C}_{il}} \in \mathbb{R}^2$ for the $l^{th}$ map point seen by the $i^{th}$ keyframe is:
\begin{equation}
    \bm{\mathrm{r}}_{\mathcal{C}_{il}}= \bm{\mathrm{z}}_{il} - \widetilde{\bm{\mathrm{z}}}_{il}
\label{visualfactor}
\end{equation}

\subsection{Plane Factor}
The x-y plane of the first wheel encoder frame $\{O_1\}$ coincides with the physical plane, so the planar measurement in section \ref{preliminaries} corresponds to that the roll, pitch angle and translation on z-axis between frame $\{O_1\}$ and $\{O_k\}$ should be close to zero. Therefore, we express the plane factor $\bm{\mathrm{r}}_\mathrm{pl} \in \mathbb{R}^3$ as:
\begin{equation}
   \bm{\mathrm{r}}_\mathrm{pl} =  \left[ \begin{array}{c} {\left[ \begin{array}{cc}  \bm{\mathrm{e}}_{1} & \bm{\mathrm{e}}_{2} \end{array} \right]}^\mathrm{T}  \bm{\mathrm{R}}^{O_k}_{W}{\bm{\mathrm{R}}^{O_1}_{W}}^\mathrm{T} \bm{\mathrm{e}}_{3} \\ \bm{\mathrm{e}}_{3}^\mathrm{T} (-\bm{\mathrm{R}}^{O_1}_{W} {\bm{\mathrm{R}}^{O_k}_{W}}^\mathrm{T} \bm{\mathrm{p}}^{O_k}_{W} + \bm{\mathrm{p}}^{O_1}_{W}) \end{array} \right] - \widetilde{\bm{\mathrm{pl}}}_k
\end{equation}

\subsection{On-Manifold Optimization}
The MAP estimate in \eqref{MAPfactor} can be written in general form on manifold $\mathcal{M}$ as:
\begin{equation}
\begin{split}
&\mathrm{F}_{mn} = {\bm{\mathrm{r}}(\bm{\mathcal{X}}_m, \bm{\mathcal{X}}_n, \bm{\mathcal{Z}}_{mn})}^\mathrm{T} \bm{\Sigma}_{mn}^{-1} \bm{\mathrm{r}}(\bm{\mathcal{X}}_m, \bm{\mathcal{X}}_n, \bm{\mathcal{Z}}_{mn}) \\
&\mathrm{F}(\bm{\mathcal{X}}) =  \sum \limits_{<m,n> \in \{\mathcal{K},\mathcal{L}\}} \mathrm{F}_{mn}\\
&\bm{\mathcal{X}}^\ast = \mathop{\mathrm{argmin}} \limits_{\bm{\mathcal{X}} \in \mathcal{M}} \ \mathrm{F}(\bm{\mathcal{X}})
\label{generalMAP}
\end{split}
\end{equation}

We use the retraction approach to solve the optimization problem on manifold. The retraction $\mathcal{R}$ is a bijective map between the tangent space and the manifold. Therefore, we can re-parameterize our problem as follows:
\begin{equation}
     \bm{\mathcal{X}}^\ast = \mathcal{R}_{\bm{\mathcal{X}}}({\delta \bm{\mathcal{X}}}^\ast) = \mathop{\mathrm{argmin}} \limits_{\delta \bm{\mathcal{X}} \in \mathbb{R}^n} \ \mathrm{F}(\mathcal{R}_{\bm{\mathcal{X}}}({\delta \bm{\mathcal{X}}}))
\end{equation}
where $\delta \bm{\mathcal{X}}$ is an element of the tangent space and the minimum dimension error representation. The objective function $\mathrm{F}(\mathcal{R}_{\bm{\mathcal{X}}}({\delta \bm{\mathcal{X}}}))$ is defined on the Euclidean space, so it is easy to compute Jacobian.

For the rigid-body transformation $\bm{\mathrm{SE}}(3)$, the retractraction at $\bm{\mathrm{T}} = [\bm{\mathrm{R}}, \bm{\mathrm{p}}]$ is:
\begin{equation}
    \mathcal{R}_{\bm{\mathrm{R}}}(\delta \bm{\varphi}) = \bm{\mathrm{R}} \mathrm{Exp}(\delta \bm{\varphi}),\ \mathcal{R}_{\bm{\mathrm{p}}} (\delta \bm{\mathrm{p}}) = \bm{\mathrm{p}} + \bm{\mathrm{R}} \delta \bm{\mathrm{p}}
\end{equation}
where $\delta \bm{\varphi} \in \mathbb{R}^3$, $\delta \bm{\mathrm{p}} \in \mathbb{R}^3$.

However, since the gyro bias and position of map points are already in a vector space, the corresponding retraction at $\bm{\mathrm{b}}_g$ and  $\bm{\mathrm{f}}^{\omega}$ are:
\begin{equation}
    \mathcal{R}_{\bm{\mathrm{b}}_g}(\delta \bm{\mathrm{b}}_g) =  \bm{\mathrm{b}}_g + \delta \bm{\mathrm{b}}_g , \mathcal{R}_{\bm{\mathrm{f}}^{\omega}}(\delta \bm{\mathrm{f}}^{\omega}) =  \bm{\mathrm{f}}^{\omega} + \delta \bm{\mathrm{f}}^{\omega}
\end{equation}
where $\delta \bm{\mathrm{b}}_g \in \mathbb{R}^3$ and $\delta \bm{\mathrm{f}}^{\omega} \in \mathbb{R}^3$.

We adopt the Gauss-Newton algorithm to solve \eqref{generalMAP} since a good initial guess $\breve{\bm{\mathcal{X}}}$ can be obtained. Firstly, we linearize each error function in \eqref{generalMAP} with respect to $\delta \bm{\mathcal{X}}$ by its first order Taylor expansion around the current initial guess  $\breve{\bm{\mathcal{X}}}$:

\begin{equation}
\begin{split}
\bm{\mathrm{r}} \left(\mathcal{R}_{\breve{\bm{\mathcal{X}}}_m}(\delta \bm{\mathcal{X}}_m), \mathcal{R}_{\breve{\bm{\mathcal{X}}}_n}(\delta \bm{\mathcal{X}}_n), \bm{\mathcal{Z}}_{mn} \right) & = \bm{\mathrm{r}}_{mn} \left( \mathcal{R}_{\breve{\bm{\mathcal{X}}}} (\delta \bm{\mathcal{X}}) \right) \\
&= \bm{\mathrm{r}}_{mn} + \bm{\mathrm{J}}_{mn} \delta \bm{\mathcal{X}}
\label{linearize}
\end{split}
\end{equation}
where, $\bm{\mathrm{J}}_{mn}$ is the jacobian of $\bm{\mathrm{r}}_{mn} \left( \mathcal{R}_{\breve{\bm{\mathcal{X}}}} (\delta \bm{\mathcal{X}}) \right)$ with respect to $\delta \bm{\mathcal{X}}$, which is computed in $\breve{\bm{\mathcal{X}}}$, and $\bm{\mathrm{r}}_{mn} = \bm{\mathrm{r}}_{mn}(\breve{\bm{\mathcal{X}}}) = \bm{\mathrm{r}}(\breve{\bm{\mathcal{X}}}_m, \breve{\bm{\mathcal{X}}}_n, \bm{\mathcal{Z}}_{mn})$. Substituting \eqref{linearize} to the each error term $\mathrm{F}_{mn}$ of \eqref{generalMAP}, we obtain:

\begin{equation}
\begin{split}
&\mathrm{F}\left( \mathcal{R}_{\breve{\bm{\mathcal{X}}}} (\delta \bm{\mathcal{X}}) \right)
= \sum \limits_{<m,n> \in \{\mathcal{K},\mathcal{L}\}} \mathrm{F}_{mn}\left( \mathcal{R}_{\breve{\bm{\mathcal{X}}}} (\delta \bm{\mathcal{X}}) \right)  \\
&= \sum \limits_{<m,n> \in \{\mathcal{K},\mathcal{L}\}} {(\bm{\mathrm{r}}_{mn} + \bm{\mathrm{J}}_{mn} \delta \bm{\mathcal{X}})}^\mathrm{T} \bm{\Sigma}_{mn}^{-1} (\bm{\mathrm{r}}_{mn} + \bm{\mathrm{J}}_{mn} \delta \bm{\mathcal{X}}) \\
&= \sum \limits_{<m,n> \in \{\mathcal{K},\mathcal{L}\}} \bm{\mathrm{r}}_{mn}^\mathrm{T} \bm{\Sigma}_{mn}^{-1} \bm{\mathrm{r}}_{mn} + 2 \bm{\mathrm{r}}_{mn}^\mathrm{T} \bm{\Sigma}_{mn}^{-1} \bm{\mathrm{J}}_{mn} \delta \bm{\mathcal{X}} \\
& \ \ \ \ \ \ \ \ \ \ \ \ \ \ \ \ \ \ \ \ \ \ \ \ \ \ \ \ \ \ \ \ \ \ \ \ + \delta \bm{\mathcal{X}}^\mathrm{T} \bm{\mathrm{J}}_{mn}^\mathrm{T} \bm{\Sigma}_{mn}^{-1} \bm{\mathrm{J}}_{mn} \delta \bm{\mathcal{X}}\\
&= \bm{\mathrm{r}}^\mathrm{T} {\bm{\Sigma}}^{-1} \bm{\mathrm{r}} + 2 \bm{\mathrm{r}}^\mathrm{T} {\bm{\Sigma}}^{-1} \bm{\mathrm{J}} \delta \bm{\mathcal{X}} + \delta \bm{\mathcal{X}}^\mathrm{T} \bm{\mathrm{J}}^\mathrm{T} {\bm{\Sigma}}^{-1} \bm{\mathrm{J}} \delta \bm{\mathcal{X}}
\end{split}
\end{equation}
where $\bm{\mathrm{r}}$, $\bm{\Sigma}^{-1}$, $\bm{\mathrm{J}}$ are formed by stacking $\bm{\mathrm{r}}_{mn}$, $\bm{\Sigma}_{mn}^{-1}$, $\bm{\mathrm{J}}_{mn}$ respectively. Then we take the derivative of $\mathrm{F}\left( \mathcal{R}_{\breve{\bm{\mathcal{X}}}} (\delta \bm{\mathcal{X}}) \right)$ with respect to $\delta \bm{\mathcal{X}}$ and set the derivative to zero, which leads to the following linear system:
\begin{equation}
    \bm{\mathrm{J}}^\mathrm{T} {\bm{\Sigma}}^{-1} \bm{\mathrm{J}} \Delta\bm{\mathcal{X}}^\ast = -\bm{\mathrm{r}}^\mathrm{T} {\bm{\Sigma}}^{-1} \bm{\mathrm{J}}
\label{deltacompute}
\end{equation}

Finally, the state is updated by adding the increment ${\delta \bm{\mathcal{X}}}^\ast$ to the initial guess $\breve{\bm{\mathcal{X}}}$:
\begin{equation}
    \bm{\mathcal{X}} ^\ast = \mathcal{R}_{\breve{\bm{\mathcal{X}}}} ({\delta \bm{\mathcal{X}}}^\ast)\
\label{update}
\end{equation}

Following the scheme of the Gauss-Newton algorithm, we solve \eqref{generalMAP} by iterating linearization in \eqref{linearize}, the computation of increments in \eqref{deltacompute}, and the state update in \eqref{update} until a given termination criterion is met. Moreover, the previous solution is used as the initial guess for each iteration.

\section{monocular visual-odometric SLAM}
Our monocular visual-odometric SLAM system is inspired by the ORB-SLAM \cite{ORBSLAM} and visual-inertial ORB-SLAM \cite{ORBVIO} methods. Fig. \ref{FLOWCHART} shows an overview of our system. In this section, we detail the main changes of our visual-odometric SLAM system with respect to the referenced system.

\subsection{Map Initialization}
The map initialization is in charge of constructing an initial set of map points by using the visual and odometer data. Firstly, we extract ORB features in the current frame $k$ and search for feature correspondences with reference frame $r$. If there are sufficient feature matches, we perform the next step, else we set the current frame as reference frame. The second step is to check the parallax of each correspondence and pick out a set of feature matches $\mathcal{F}$ that have sufficient parallax. When the size of $\mathcal{F}$ is greater than a threshold, we use the odometer measurements to compute the relative transformation between two frames, and triangulate the matched features $\mathcal{F}$. Finally, if the size of the successfully created map points is greater than a threashold, a global BA that minimizes all reprojection error, odometer error and plane error in the initial map is applied to refine the initial map.

\begin{figure*}[!t]
\centering
\includegraphics[width=7in]{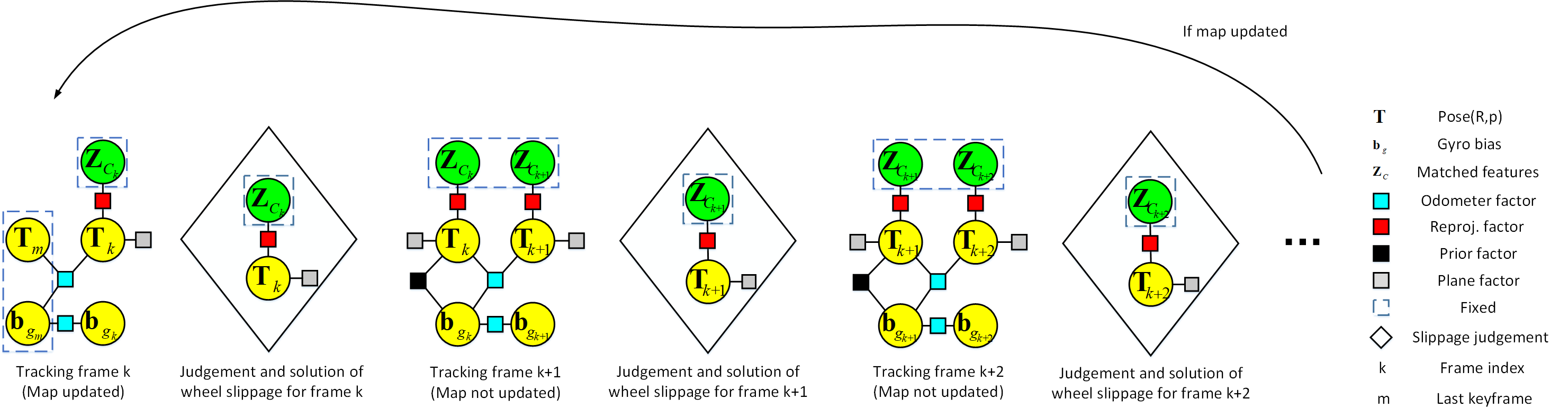}
\caption{Evolution of the factor graph in the tracking thread when previous visual tracking is successful. If map is updated, we optimize the state of frame $k$ by connecting an odometer factor to last keyframe $m$. If map is not changed, state of both last frame $k-1$ and current frame $k$ are jointly optimized by linking an odometer factor between them and adding a prior factor to last frame $k-1$. The prior for last frame $k-1$ is obtained from the previous optimization. At the end of each joint optimization, we judge whether the wheel slippage has occurred through the optimized result. If wheel slippage is detected, the pose of frame $k$ is re-optimized using factor graph in the diamond.}
\label{tracking}
\end{figure*}

\subsection{Tracking when Previous Visual Tracking is Successful} \label{visualsuccess}
Once the initial pose of current frame is predicted using the odometer measurements, the map points in the local map are projected into the current frame and matched with the keypoints extracted from the current frame. Then the pose of current frame is optimized by minimizing the corresponding energy function. Depending on whether the map in back-end is updated, the pose prediction and optimization methods are different, which will be described in detail below. In addition, we provide a detection strategy and solution for wheel slippage. The tracking mechanism is illustrated in Fig. \ref{tracking}.

\subsubsection{Tracking when Map Updated}
When tracking is performed just after a map update in the back-end, we firstly compute the preintegrated odometer measurement between current frame $k$ and last keyframe $m$. Then the computed relative transformation $\bm{\mathrm{T}}^{O_k}_{O_m}$ is combined with the optimized pose of last keyframe to predict the initial pose of the current frame. The reason for this state prediction is that the pose estimate of the last keyframe is accurate enough after performing a local or global BA in the back-end. Finally, the state of the current frame $k$ is optimized by minimizing the following energy function:
\begin{equation}
\begin{split}
&\bm{\gamma} = \{ \bm{x}_k \}\\
&\bm{\gamma} ^\ast = \mathop{\mathrm{argmin}} \limits_{\bm{\gamma}} \left( \sum \limits_{l \in \mathcal{Z}_{C_k}} {\|\bm{\mathrm{r}}_{\mathcal{C}_{kl}}}\|^2_{{\bm{\Sigma}}_{\mathcal{C}_{kl}}} +\|{\bm{\mathrm{r}}_{\mathcal{O}_{mk}}}\|^2_{{\bm{\Sigma}}_{\mathcal{O}_{mk}}} + \|{\bm{\mathrm{r}}_{\mathrm{pl}_{k}}}\|^2_{{\bm{\Sigma}}_\mathrm{pl}} \right)
\end{split}
\end{equation}

After the optimization, the resulting estimation and Hessian matrix are served as a prior for next optimization.

\subsubsection{Tracking when no Map Updated}
When the map is not changed in the back-end, we compute the preintegrated odometer measurement between current frame $k$ and last frame $k-1$, and predict the initial pose of current frame $k$ by integrating the relative transformation $\bm{\mathrm{T}}^{O_k}_{O_{k-1}}$ to the pose of last frame. Then, we optimize the pose of current frame $k$ by performing the nonlinear optimization that minimizing the following objective function:
\begin{equation}
\begin{split}
&\bm{\gamma} = \{ \bm{x}_{k-1}, \bm{x}_k \}\\
&\bm{\gamma} ^\ast = \mathop{\mathrm{argmin}} \limits_{\bm{\gamma}} ( \sum \limits_{l \in \mathcal{Z}_{C_{k-1}}} {\|\bm{\mathrm{r}}_{\mathcal{C}_{k-1l}}}\|^2_{{\bm{\Sigma}}_{\mathcal{C}_{k-1l}}}  + \sum \limits_{n \in \mathcal{Z}_{C_k}} {\|\bm{\mathrm{r}}_{\mathcal{C}_{kn}}}\|^2_{{\bm{\Sigma}}_{\mathcal{C}_{kn}}} \\
& \ \ \ \ \ \ \ \ \ \ \ \ \ \ \ \ \ \ \ \ + \|{\bm{\mathrm{r}}_{\mathcal{O}_{k-1k}}}\|^2_{{\bm{\Sigma}}_{\mathcal{O}_{k-1k}}} + \|\bm{\mathrm{r}}_{0_{k-1}}\|^2_{{\bm{\Sigma}}_{0_{k-1}}}\\
& \ \ \ \ \ \ \ \ \ \ \ \ \ \ \ \ \ \ \ \ + \|{\bm{\mathrm{r}}_{\mathrm{pl}_{k-1}}}\|^2_{{\bm{\Sigma}}_\mathrm{pl}} + \|{\bm{\mathrm{r}}_{\mathrm{pl}_{k}}}\|^2_{{\bm{\Sigma}}_\mathrm{pl}} )
\end{split}
\end{equation}
where the residual $\textbf{r}_{{0}_{k-1}} = {\{ \bm{\mathrm{r}}_{\bm{\mathrm{R}}_{k-1}}^\mathrm{T}, \bm{\mathrm{r}}_{\bm{\mathrm{p}}_{k-1}}^\mathrm{T}, \bm{\mathrm{r}}_{\bm{\mathrm{b}}_{k-1}}^\mathrm{T} \}}^\mathrm{T} \in \mathbb{R}^9$ is a prior error term of last frame:
\begin{equation}
\begin{split}
&{\bm{\mathrm{r}}_{\bm{\mathrm{R}}_{k-1}}} = \mathrm{Log} \left(\bm{\mathrm{R}}^{O_{k-1}}_W * {\widetilde{\bm{\mathrm{R}}}^{O_{k-1}}_W} \right)\\
&{\bm{\mathrm{r}}_{\bm{\mathrm{p}}_{k-1}}} = \bm{\mathrm{p}}^{O_{k-1}}_\omega - \widetilde{\bm{\mathrm{p}}}^{O_{k-1}}_\omega\\
&{\bm{\mathrm{r}}_{\bm{\mathrm{b}}_{k-1}}} = \bm{\mathrm{b}}_{g_{k-1}} - \widetilde{\bm{\mathrm{b}}}_{g_{k-1}}
\end{split}
\end{equation}
where $\widetilde{\bm{\mathrm{R}}}^{O_{k-1}}_W$, $\widetilde{\bm{\mathrm{p}}}^{O_{k-1}}_\omega$, $\widetilde{\bm{\mathrm{b}}}_{g_{k-1}}$ and ${{\bm{\Sigma}}_{0_{k-1}}}$ are the estimated states and resulting covariance matrix from previous pose optimization. The optimized result is also served as a prior for next optimization.

\subsubsection{Detecting and Solving Wheel Slippage}
Wheel encoder is an ambivalent sensor, it provides a precise and stable relative transformation at most of the time, but it can also deliver very faulty data when the robot experiences slippage. If we perform visual-odometric joint optimization using this kind of faulty data, in order to simultaneously satisfy the constraints of both odometer measurements with slippage and visual measurements, the optimization will lead to a false estimate. Therefore, we provide a strategy to detect and solve this case. We think the current frame $k$ experienced a slippage if the above optimization makes more than half of the original matched features become outliers. Once the wheel slippage is detected, we set slippage flag to current frame and reset the initial pose of current frame $k$ as the pose of last frame $k-1$. Then we re-project the map points in the local map and re-match with features on the current frame. Finally, the state of current frame is optimized by only using those matched features:
\begin{equation}
\begin{split}
&\bm{\gamma} = \{ \bm{x}_k \}\\
&\bm{\gamma} ^\ast = \mathop{\mathrm{argmin}} \limits_{\bm{\gamma}} \left( \sum \limits_{l \in \mathcal{Z}_{C_k}} \bm{\mathrm{r}}_{\mathcal{C}_{kl} } + \|{\bm{\mathrm{r}}_{\mathrm{pl}_{k}}}\|^2_{{\bm{\Sigma}}_\mathrm{pl}} \right)
\end{split}
\end{equation}

After the optimization, the resulting estimate and Hessian matrix of current frame are served as a prior for next optimization.

\subsection{Tracking when Previous Visual Tracking is Lost}\label{trackingwhenpreviousvisualtrackingislost}
If visual information is not available in current frame, only odometer measurements can be used to compute the pose of the frame. So in order to obtain more accurate pose estimate, we should make the visual information available as early as possible. 

Supposing the previous visual tracking is lost, then one of the three cases will happen for the current frame: (1) the robot revisits to an already reconstructed area; (2) the robot visits to a new environment where sufficient map points are newly constructed; (3) the visual features are still unavailable wherever the robot is. For these different situations, we perform different strategies to estimate the pose of the current frame. For case 1, a global relocalization method as done in \cite{ORBSLAM}, i.e. using DBOW\cite{DBOW2} and PnP algorithm\cite{PNP}, is performed to compute the pose of the current frame and render the visual information available. For case 2, we firstly use the odometer measurements to predict the initial pose of current frame, then project map points seen by last keyframe to the current frame and optimize the pose of current frame using those matched features. For case 3, we use the odometer measurements to compute the pose of the current frame.

When enough features are extracted from the current frame after visual tracking is lost, we firstly think the robot may returned to an already reconstructed environment, therefore perform the global relocaliation method(solution for case 1). However, if the relocalization continuously fails until the second keyframe with enough features is selected to enable the reconstruction of the new map, we think the robot entered into a new environment, thereby the localization in newly constructed map is performed as solution for case 2. We deem the visual information becomes available for motion tracking of current frame when the camera pose is supported by enough matched features. So if the computed pose in case 1 and case 2 is not supported by enough matched features or fewer features are extracted from the current frame, we think the visual information is still unavailable for motion tracking of the current frame and set the pose of current frame according to the solution for case 3.

\subsection{Keyframe Decision}
If the visual tracking is successful, we have two criteria for keyframe selection: (1) current frame tracks less than 50$\%$ features than last keyframe; (2) Local BA is finished in the back-end. These criteria insert keyframes as many as possible to make visual tracking and mapping to work all the time, thereby ensure a good performance of the system.

In addition, if visual tracking is lost, we insert a keyframe to the back-end when one of the following conditions is satisfied: (1) The traveled distance from the last keyframe is large than a threshold; (2) The relative rotation angle from the last keyframe is beyond a threshold; (3) Local BA is finished in the local mapping thread. These conditions ensure that when the previous map is not available and the robot enters into a new environment where there are enough features, the system can still build new map that are consistent with the previous map.

\subsection{Back-End}
The back-end includes the local mapping thread and the loop closing thread. The local mapping thread aims to construct the new map points of the environment and optimize the local map. When new keyframe $k$ is inserted to local mapping thread, we make small changes in convisibility graph update and local BA with respect to paper \cite{ORBVIO}. If the visual tracking of new keyframe $k$ is lost, we update the covisibility graph by adding a new node for keyframe $k$ and an edge connected with the last keyframe to ensure the ability to build new map. In addition, visual-odometric local BA is performed to optimize the last N keyframes(local window) and all points seen by those N keyframes, which is achieved by minimizing the cost function \eqref{MAPfactor} in the window. One thing to note is that the odometer constraint linking to the previous keyframe is only constructed for those keyframes without the slippage flag. The loop closing thread is in charge of eliminating the accumulated drift when returning to an already reconstructed area, it is implemented in the same way as paper \cite{ORBVIO}.

\section{EXPERIMENTS}
In the following, we perform a number of experiments to evaluate the proposed approach both qualitatively and quantitatively. Firstly, we perform qualitative and quantitative analysis of our algorithm to show the accuracy of our system in Section \ref{algorithmevaluation}. Then the validity of the proposed strategy for detecting and solving the wheel slippage is demonstrated in Section \ref{robustnessvali}. Finally in Section \ref{visuallostvali}, we test the tracking performance of the algorithm when the previous visual tracking is lost. The experiments are performed on a laptop with Intel Core i5 2.2GHz CPU and an 8GB RAM, and the corresponding videos are available at: \href{https://youtu.be/EaDTC92hQpc}{https://youtu.be/EaDTC92hQpc}. In addition, our system is able to work robustly in raspberry pi platform that has a Quad Core 1.2GHz Broadcom BCM2837 64bit CPU and 1GB RAM, at the processing frequency of 5Hz.

\begin{figure}[!h]
\centering
\includegraphics[width=3.5in]{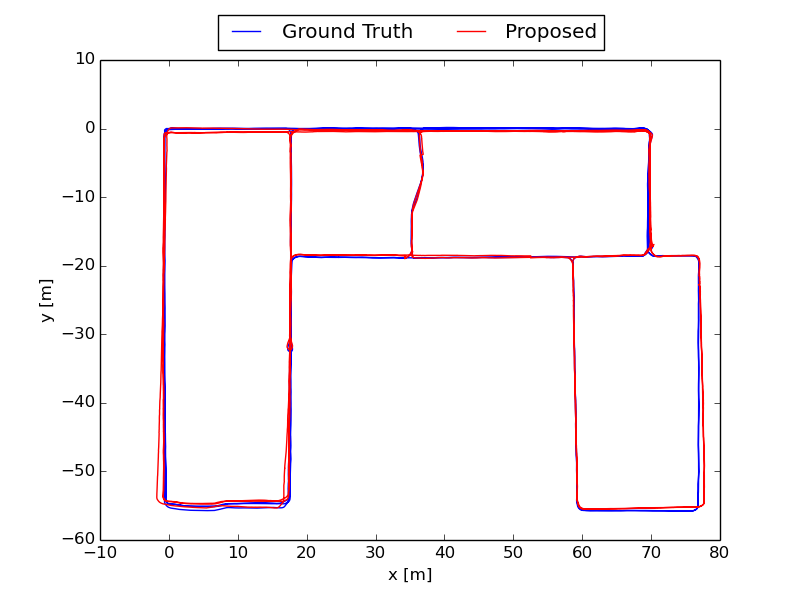}
\caption{Comparison between the estimated trajectory and the ground truth.}
\label{dstrajectory}
\end{figure}

\subsection{Algorithm Evaluation}\label{algorithmevaluation}
We evaluate the accuracy of the proposed algorithm in a dataset provided by the author of \cite{VINSWHEEL}. The dataset is recorded by a Pioneer 3 DX robot with a Project Tango, and provides 640 $\times$ 480 grayscale images at 30 Hz, the inertial measurements at 100Hz and wheel-encoder measurements at 10 Hz. In addition, the dataset also provides the ground truth which is computed from the batch least squares offline using all available visual, inertial and wheel measurements.

We process images at the frequency of 10 Hz, qualitative comparison of the estimated trajectory and the ground truth is shown in Fig. \ref{dstrajectory}. The estimated trajectory and the ground truth are aligned in closed form using the method of Horn\cite{Horn}. We can qualitatively compare our estimated trajectory with the result provided by the approach of Wu. et. al \cite{VINSWHEEL} in their figure 6. It is clear that our algorithm produces more accurate trajectory estimate, which is achieved by executing the complete visual-odometric tracking strategies, performing the local BA to optimize the local map and closing loop to eliminate the accumulated error when returning to an already mapped area. Quantitatively, the sequence is 1080m long, and the positioning Root Mean Square Error(RMSE) of our algorithm is 0.606m, it is the 0.056\% of the total traveled distance with comparison to 0.25\% of the approach \cite{VINSWHEEL}.

\subsection{Demonstration of Robustness to Wheel Slippage} \label{robustnessvali}
In the following experiments, we use data recorded from a DIY robot with a OV7251 camera mounted on it to look upward for visual sensing. The sensor suite provides the 640 $\times$ 480 grayscale images at the frequency of 30Hz, the wheel-odometer and gyroscope measurements at 50 Hz. Since there is no ground truth available, we only do qualitatively analysis.

\begin{figure*}[!h]
\centering
\subfloat[wheel begins to slip]{
\label{wheelslippage1im:a}
\includegraphics[width=2.1in]{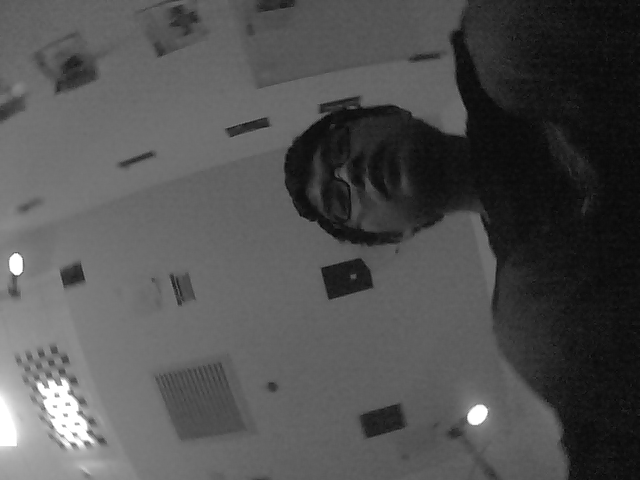}
}
\subfloat[wheel slippage is over]{
\label{wheelslippage1im:b}
\includegraphics[width=2.1in]{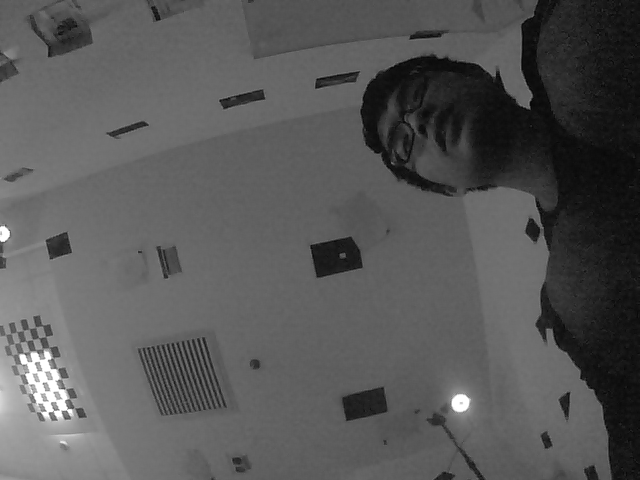}
}
\subfloat[end of sequence]{
\label{wheelslippage1im:c}
\includegraphics[width=2.1in]{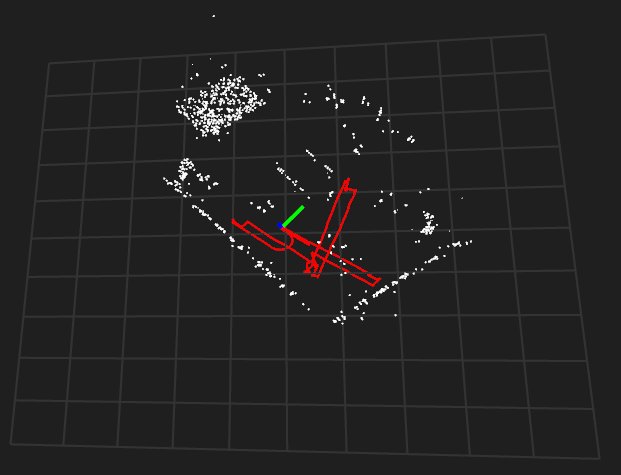}
}
\caption{Sample images when the platform begins to experience wheel slippage and wheel slippage is over, and the finally reconstructed 3D map at the end of the sequence.}
\label{wheelslippage1im}
\end{figure*}

\begin{figure*}[!h]
\centering
\subfloat[wheel begins to slip]{
\label{wheelslippage1trajectory:a}
\includegraphics[width=2.2in]{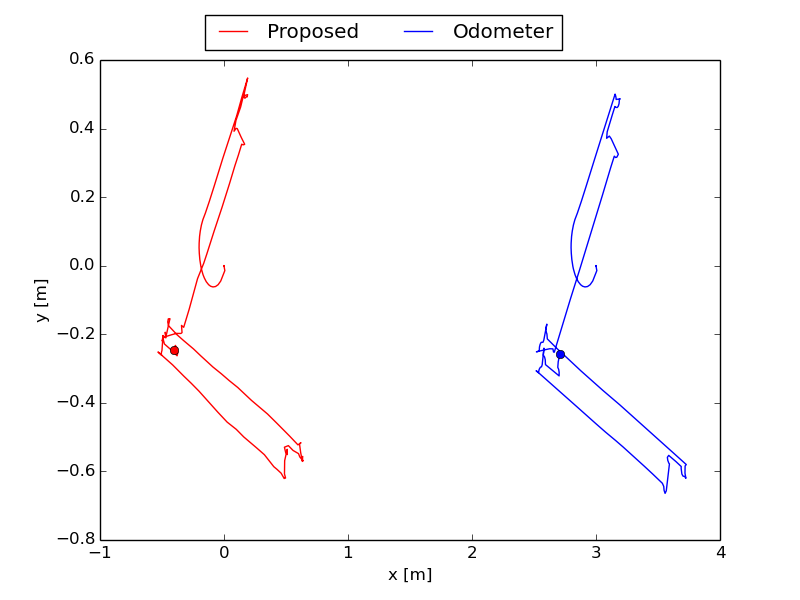}
}
\subfloat[wheel slippage is over]{
\label{wheelslippage1trajectory:b}
\includegraphics[width=2.2in]{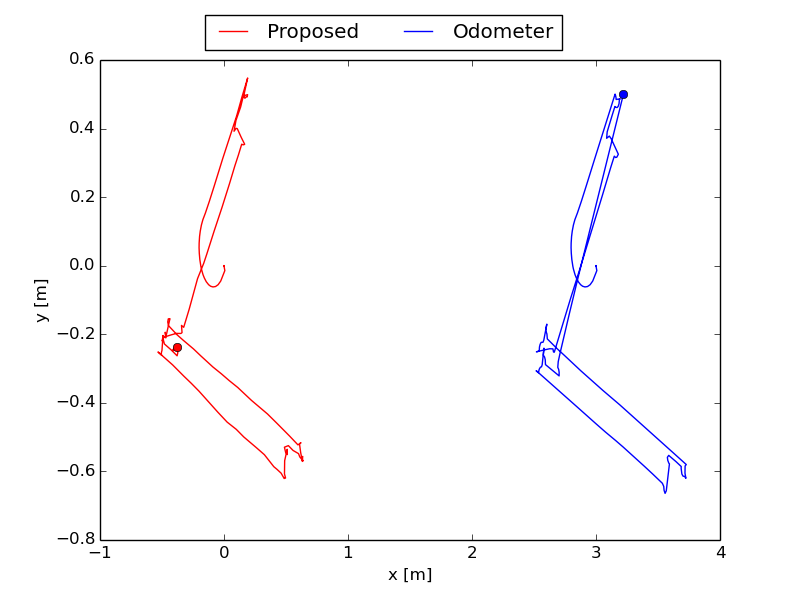}
}
\subfloat[end of sequence]{
\label{wheelslippage1trajectory:c}
\includegraphics[width=2.2in]{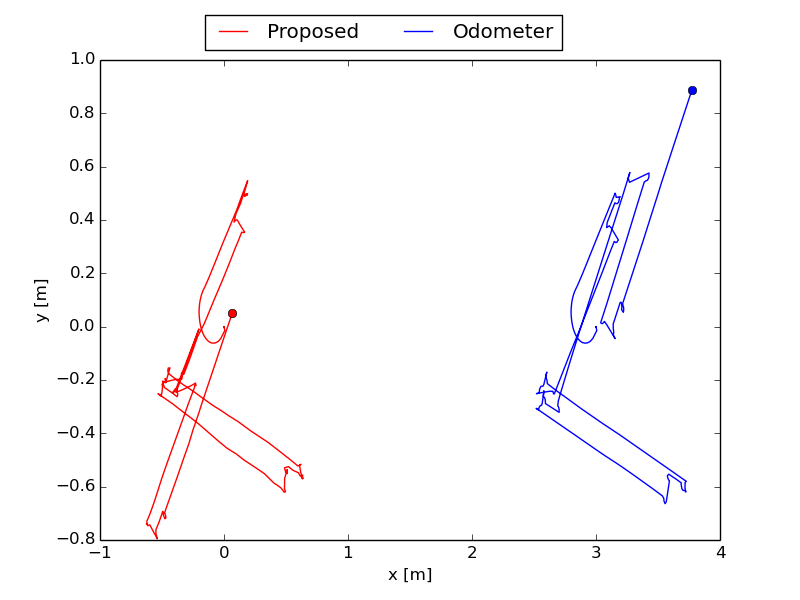}
}
\caption{The estimated trajectories from beginning to some critical moments.}
\label{wheelslippage1trajectory}
\end{figure*}

\begin{figure*}[!h]
\centering
\subfloat[wheel begins to slip]{
\label{wheelslippage2im:a}
\includegraphics[width=2.1in]{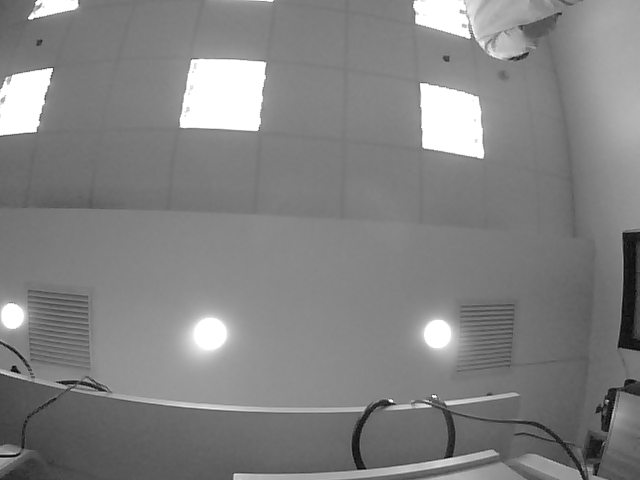}
}
\subfloat[wheel slippage is over]{
\label{wheelslippage2im:b}
\includegraphics[width=2.1in]{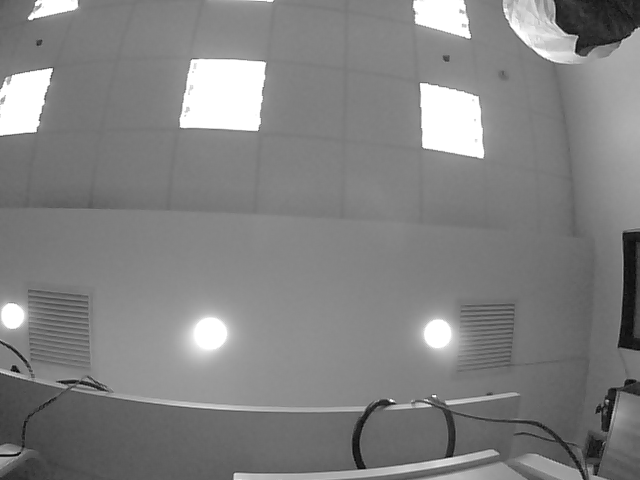}
}
\subfloat[end of sequence]{
\label{wheelslippage2im:c}
\includegraphics[width=2in]{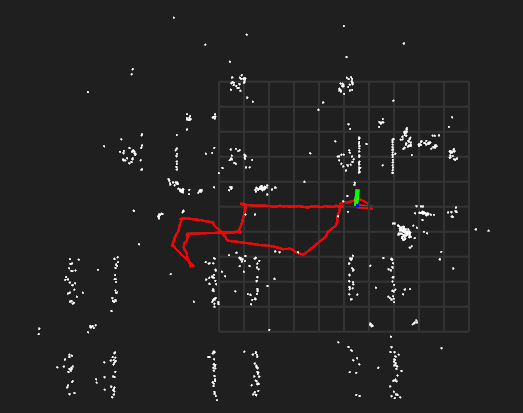}
}
\caption{Sample images when the platform begins to experience wheel slippage and wheel slippage is over, and the finally reconstructed 3D map at the end of the sequence.}
\label{wheelslippage2im}
\end{figure*}

\begin{figure*}[!h]
\centering
\subfloat[wheel begins to slip]{
\label{wheelslippage2trajectory:a}
\includegraphics[width=2.2in]{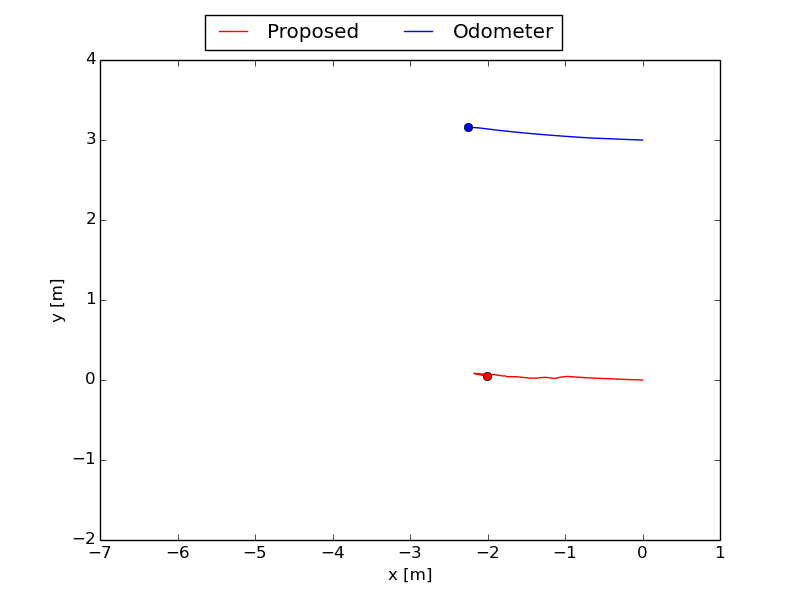}
}
\subfloat[wheel slippage is over]{
\label{wheelslippage2trajectory:b}
\includegraphics[width=2.2in]{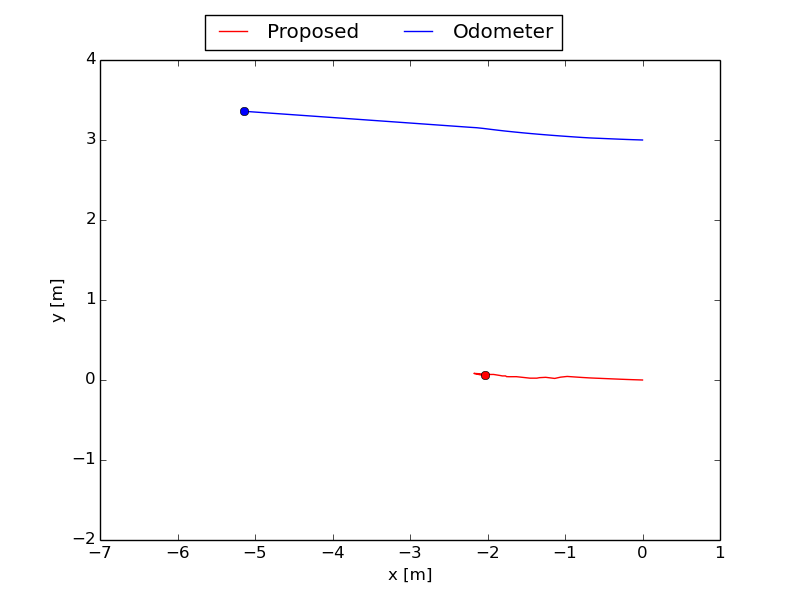}
}
\subfloat[end of sequence]{
\label{wheelslippage2trajectory:c}
\includegraphics[width=2.2in]{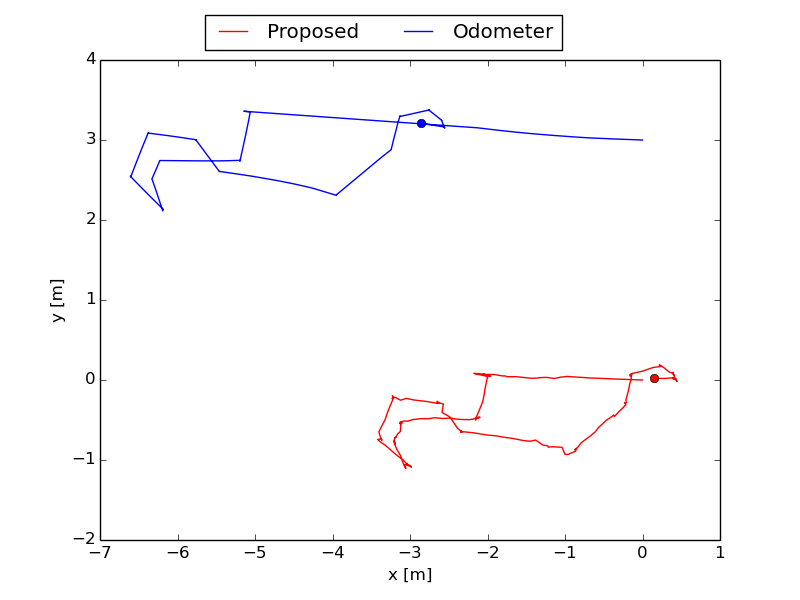}
}
\caption{The estimated trajectories from beginning to some critical moments.}
\label{wheelslippage2trajectory}
\end{figure*}

The wheel slippage experiment is performed in two situations. In the first experiment, we firstly let the ground robot to walk normally, then hold the robot to make it static but the wheel is spinning, and finally let it to normally walk once again. The estimated results in some critical moments are shown in Fig. \ref{wheelslippage1im} and Fig. \ref{wheelslippage1trajectory}. Fig. \ref{wheelslippage1im:a} is the captured image at the first critical moment when the platform start to experience wheel slippage, and the trajectories estimated by our method and the odometer from the beginning to this moment are shown in Fig. \ref{wheelslippage1trajectory:a}. We can see that both methods can accurately estimate the position of the sensor suite under normal motion. The image and the estimated trajectory obtained at second moment when wheel slippage is over are given in Fig. \ref{wheelslippage1im:b} and Fig. \ref{wheelslippage1trajectory:b}. As evident, the images at first and second moments are almost the same, our method gives the very close pose for these two moments with comparison to the odometer who provides far away positions for these two moments due to the wheel slippage. Thus the validity of the proposed strategy for detecting and solving wheel slippage can be proved. The reconstructed 3D map for the sequence are shown in Fig. \ref{wheelslippage1im:c}, the map is globally consistent, which is achieved by effectively solving the problem of wheel slippage. The situation is also tested in artificial lighting and relatively low texture environment. In Fig. \ref{wheelslippage2im} and \ref{wheelslippage2trajectory}, its intermediate and final results are given, which also demonstrates the robustness of our system to the slippage of wheel encoder.

\begin{figure*}[!h]
\centering
\subfloat[platform begins to move]{
\label{moveaway2im:a}
\includegraphics[width=2.1in]{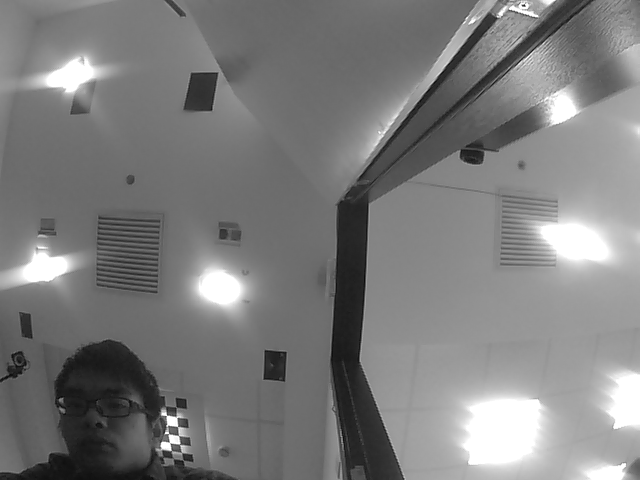}
}
\subfloat[platform has been moved]{
\label{moveaway2im:b}
\includegraphics[width=2.1in]{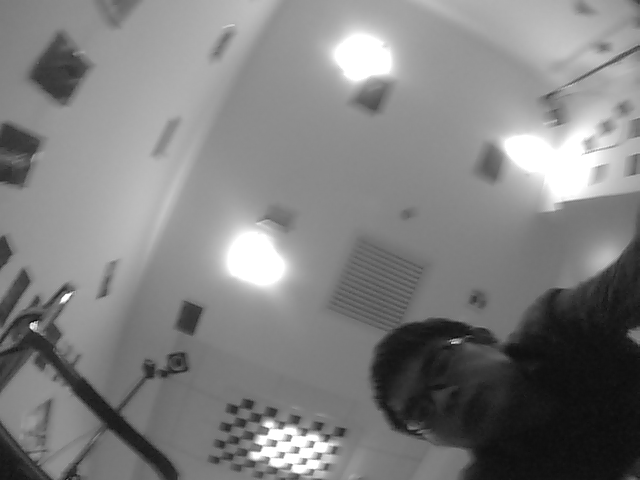}
}
\subfloat[end of sequence]{
\label{moveaway2im:c}
\includegraphics[width=2.2in]{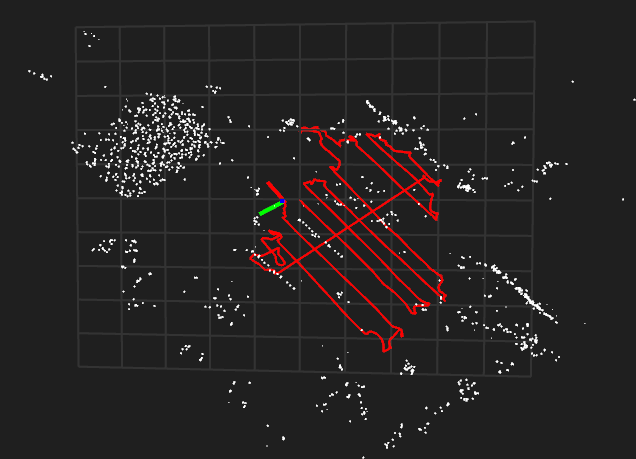}
}
\caption{Sample images when the platform begins to move and the platform has been moved to another location, and the finally reconstructed 3D map at the end of the sequence.}
\label{moveaway2im}
\end{figure*}

\begin{figure*}[!h]
\centering
\subfloat[platform begins to move]{
\label{moveaway2trajectory:a}
\includegraphics[width=2.2in]{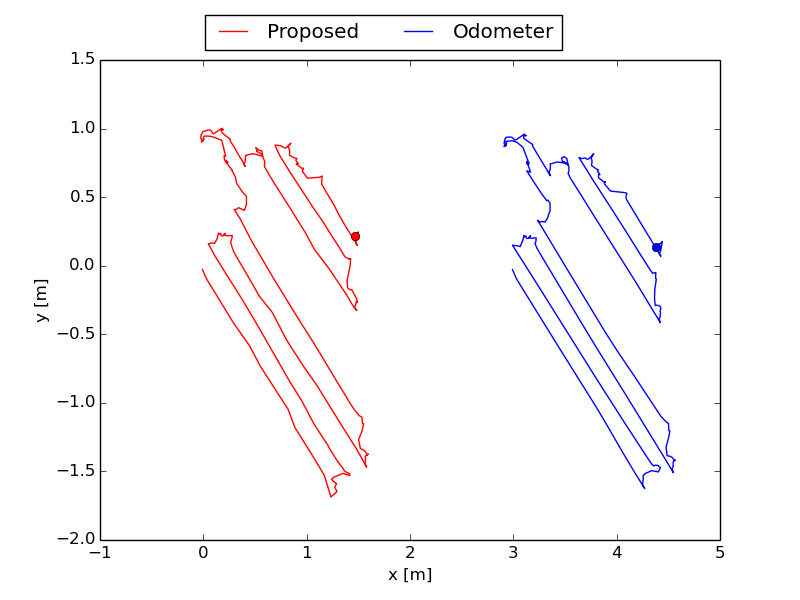}
}
\subfloat[platform has been moved]{
\label{moveaway2trajectory:b}
\includegraphics[width=2.2in]{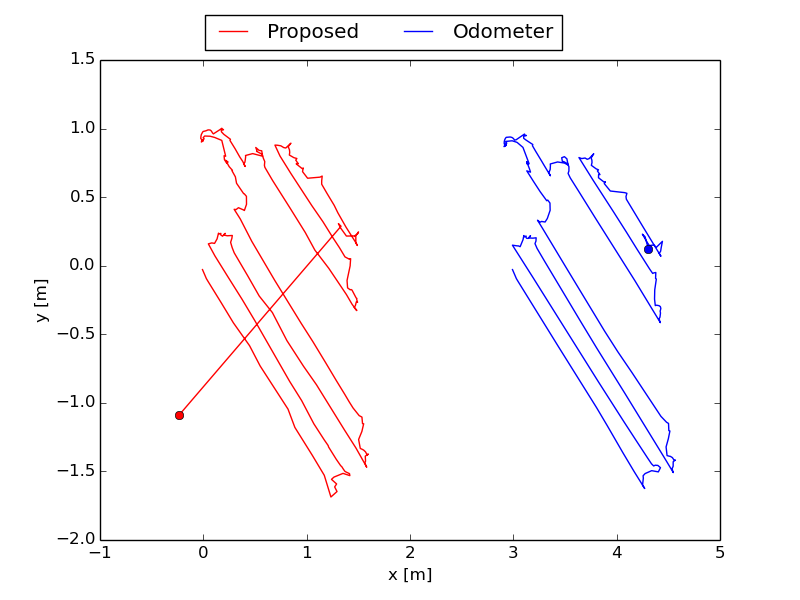}
}
\subfloat[end of sequence]{
\label{moveaway2trajectory:c}
\includegraphics[width=2.2in]{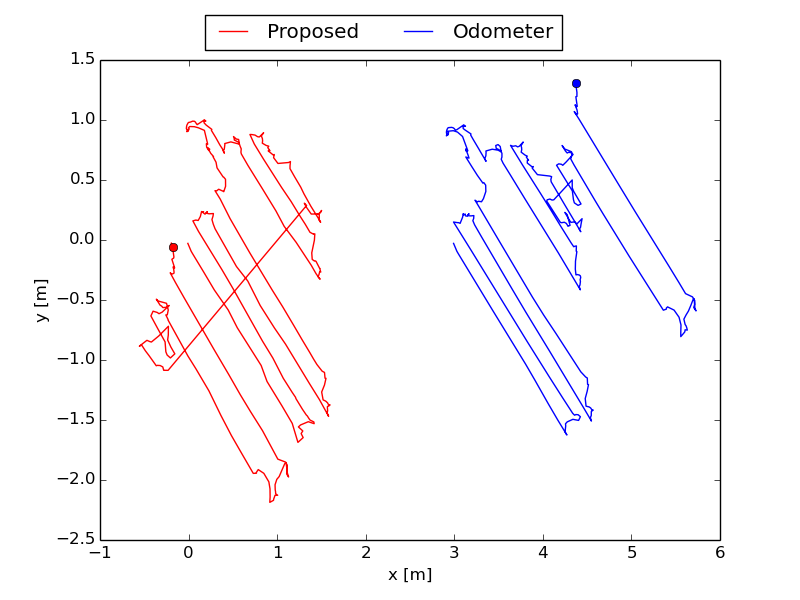}
}
\caption{The estimated trajectories from beginning to some critical moments.}
\label{moveaway2trajectory}
\end{figure*}

The second experiment is performed as follows. The sensor suite walks normally at first, then the wheel turns normally, however the platform is moved to another location artificially, and finally it normally walks once again. The test results for the second situation are shown in Fig. \ref{moveaway2im} and Fig. \ref{moveaway2trajectory}. Before the sensor is moved away, the estimated trajectories from both our method and the odometer are close to each other as shown in Fig. \ref{moveaway2trajectory:a}. Fig. \ref{moveaway2im:a} and Fig. \ref{moveaway2im:b} are the captured images at first moment when platform starts to move and at second moment when the platform has been moved to another location. Comparing to the estimated motion by the odometer, the proposed method gives precise tracking for the movement as shown in Fig. \ref{moveaway2trajectory:b}. Thereby, the performance of the proposed strategy for wheel slippage is demonstrated again.

\begin{figure*}[!t]
\centering
\subfloat[visual tracking begins to be lost]{
\label{relocim:a}
\includegraphics[width=2.22in]{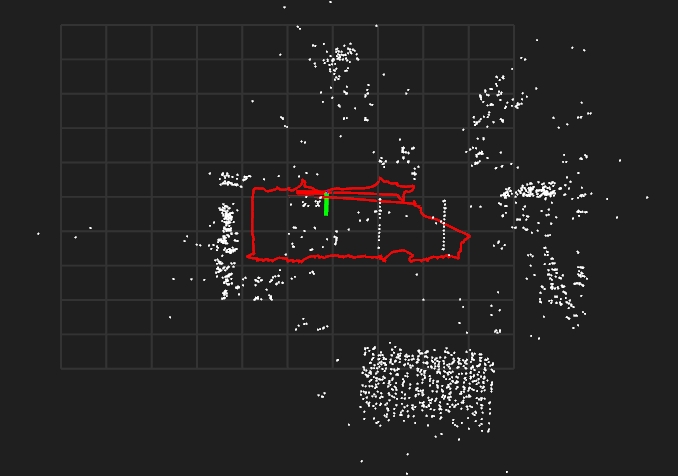}
}
\subfloat[visual tracking continues to be lost]{
\label{relocim:b}
\includegraphics[width=2.1in]{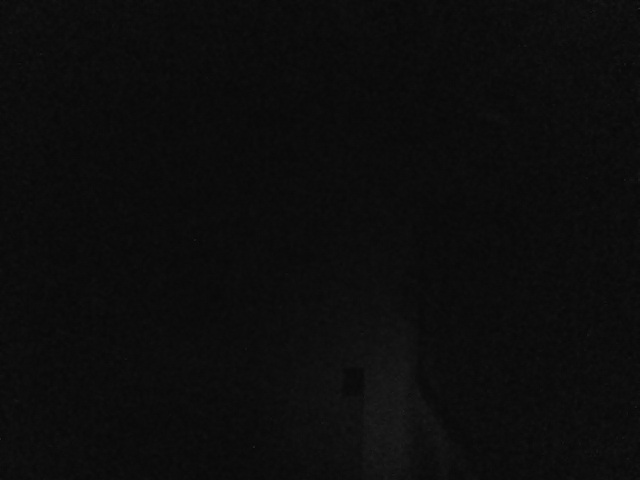}
}
\subfloat[end of sequence]{
\label{relocim:c}
\includegraphics[width=2.12in]{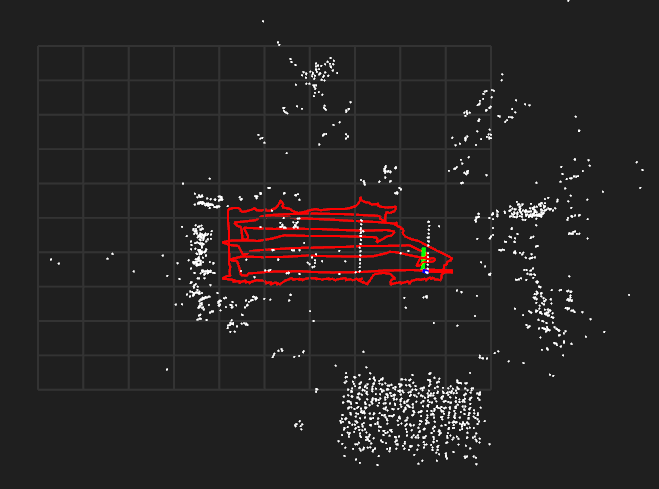}
}
\caption{Reconstructed 3D map when the visual tracking begins to be lost and at the end of the sequence, and captured image when visual tracking is lost.}
\label{relocim}
\end{figure*}

\begin{figure*}[!t]
\centering
\subfloat[visual tracking begins to be lost]{
\label{reloctra:a}
\includegraphics[width=2.2in]{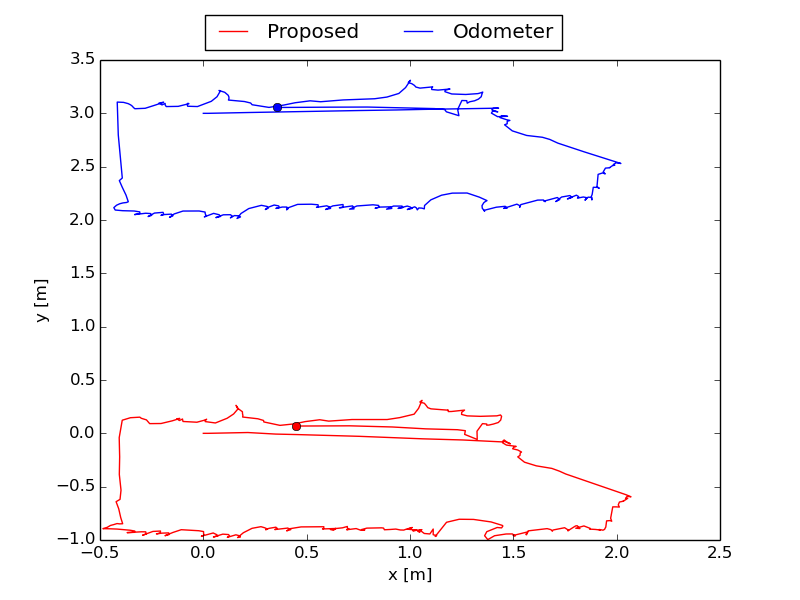}
}
\subfloat[visual tracking continues to be lost]{
\label{reloctra:b}
\includegraphics[width=2.2in]{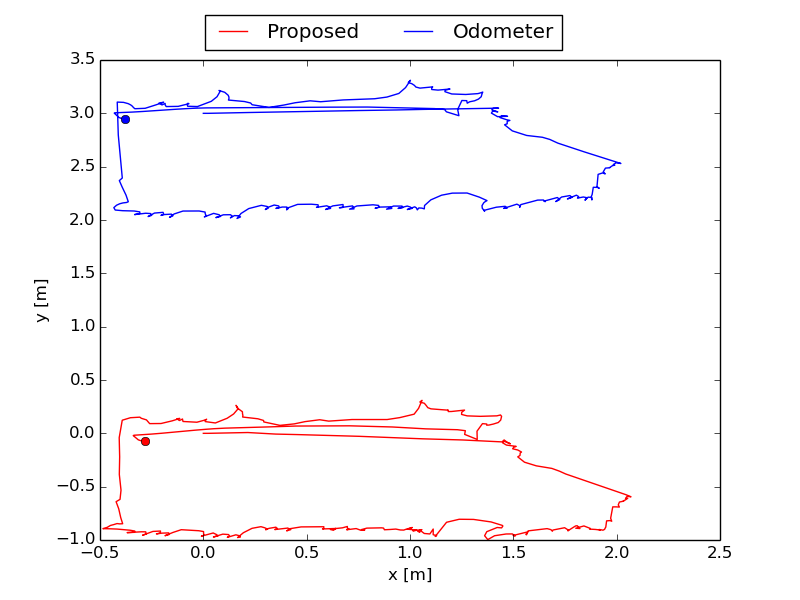}
}
\subfloat[end of sequence]{
\label{reloctra:c}
\includegraphics[width=2.2in]{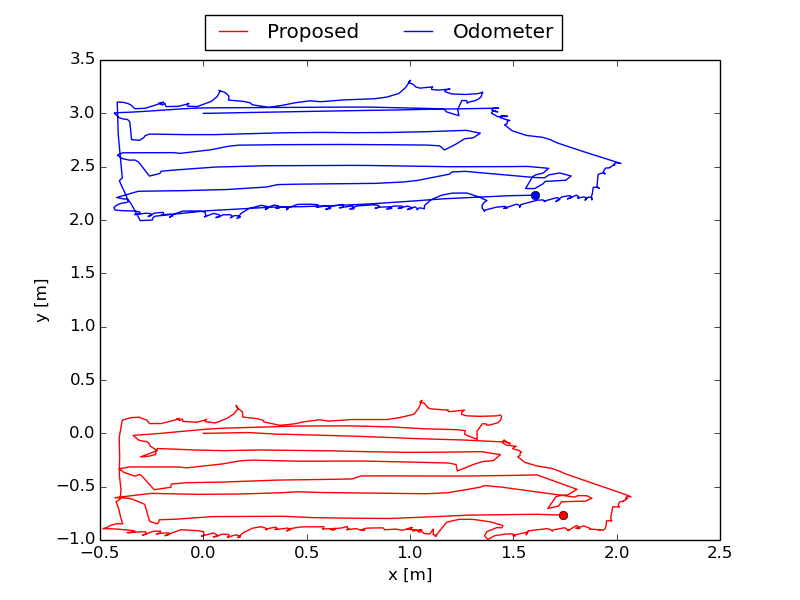}
}
\caption{The estimated trajectories from beginning to some critical moments.}
\label{reloctra}
\end{figure*}

\subsection{Demonstration of Tracking Performance when Previous Visual Tracking is Lost}\label{visuallostvali}
The tracking performance of our system when previous visual tracking is lost is tested in two sequences, sequence 1 includes the case 1 and case 3 in Section \ref{trackingwhenpreviousvisualtrackingislost} and the sequence 2 includes the case 2 and case 3 in Section \ref{trackingwhenpreviousvisualtrackingislost}. Firstly, we use sequence 1 to test the proposed solution for the case 1 and case 3, the estimated results in some critical moments are shown in Fig. \ref{relocim} and Fig. \ref{reloctra}. The robot firstly moves on areas where enough visual information is available to build a map of the environment shown in Fig. \ref{relocim:a}. Then we turn out the lights to make the visual information unavailable. The motion of robot is continuously computed in the period of visual loss as shown in Fig. \ref{reloctra:b}, which is achieved by using the odometer measurements as solution to case 3. Finally, we turn on the lights to make the robot to revisit an already reconstructed area, at which moment, the global relocalization is triggered. The reconstructed map at the end of sequence is shown in Fig. \ref{relocim:c}, it is globally consistent without closing the loop. Therefore, we can demonstrate the validity of the proposed solution for case 1 and case 3. Furthermore, we can conclude that our system is robust to visual loss, thanks to the stable measurements from the odometer.

\begin{figure*}[!t]
\centering
\subfloat[visual tracking begins to be lost]{
\label{newmapim:a}
\includegraphics[width=2.1in]{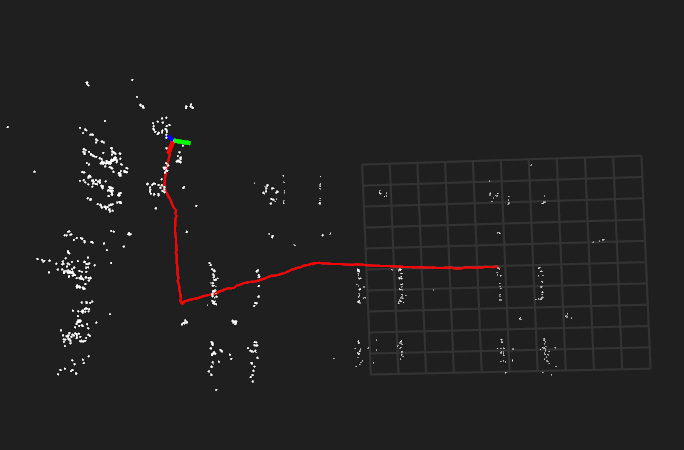}
}
\subfloat[new map after visual loss is created]{
\label{newmapim:b}
\includegraphics[width=2.1in]{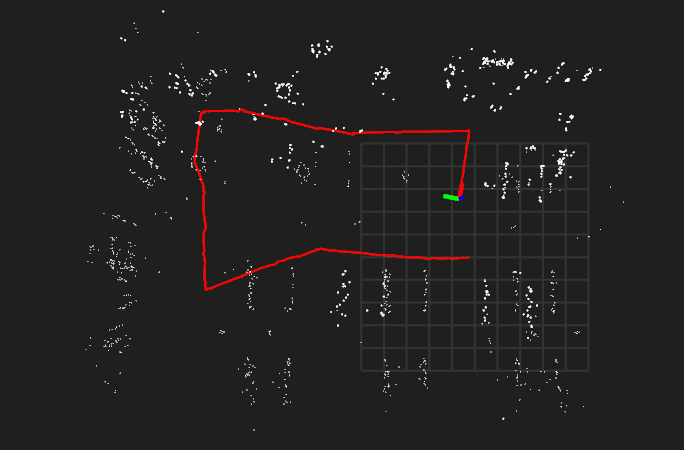}
}
\subfloat[end of sequence]{
\label{newmapim:c}
\includegraphics[width=2.11in]{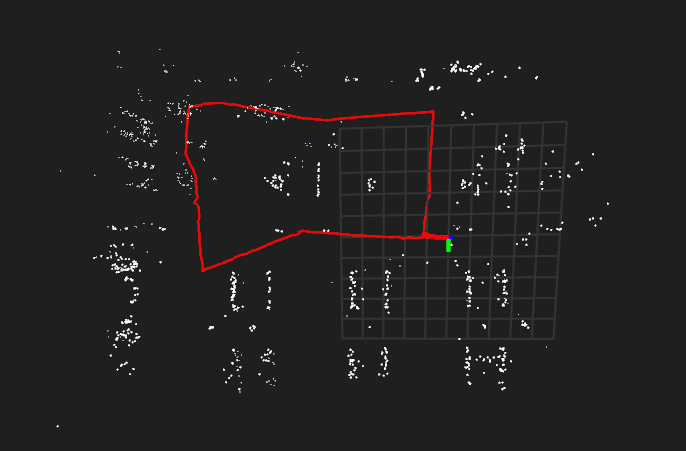}
}
\caption{Reconstructed 3D map when visual tracking begins to be lost, when the robot enters new environment where new 3D map is constructed after the visual loss and at the end of the sequence.}
\label{newmapim}
\end{figure*}

Secondly, we perform the case 2 experiment in sequence 2, the test results for the experiment are shown in Fig. \ref{newmapim}. The ground robot firstly moves on areas where enough visual information is available to build the map of the environment shown in Fig. \ref{newmapim:a}. Then the robot goes to the low texture environment, and later enters new environment where enough features are available. From Fig. \ref{newmapim:b}, we can know that new map is created when there are enough feature points in new environment, which however is not consistent with the previously reconstructed map. Finally, the robot returns to an already mapped area, this leads the system to trigger loop closure for eliminating the accumulated error, thereby globally consistent map is constructed as shown in Fig. \ref{newmapim:c}.

\section{Conclusion and future work}
In this paper, we have proposed a tightly-coupled monocular visual-odometric SLAM system. It tightly integrates the proposed odometer factor and visual factor in the optimization framework to optimally exploit the both sensor cues, which ensures the accuracy of the system. In addition, the system uses the odometer measurements to compute the motion of frame when visual information is not available, and is able to detect and reject false information from wheel encoder, thereby ensuring the robustness of the system. The experiments have domenstrated that our system can provide accurate, robust and long-term localization for the wheeled robots mostly moving on a plane.

In future work, we aim to exploit line features to improve the performance of our algorithm in environments where only fewer point features are available. In addition, the camera-to-wheel encoder calibration parameters are only known with finite precision, which can pose a bad effect on results, so we intend to estimate the extrinsic calibration parameters online and optimize this parameters by BA. Finally, we will add full IMU measurements to our system for accurate motion tracking when both visual information and wheel odometric information cannot provide valid information for localization.

\ifCLASSOPTIONcompsoc
  \section*{Acknowledgments}
\else
  \section*{Acknowledgment}
\fi

\indent \indent This paper is supported by National Science Foundation of China[grant number 61375081]; a special fund project of Harbin science and technology innovation talents research [grant number RC2013XK010002]. We thank the author of \cite{VINSWHEEL} for providing the dataset to evaluate our algorithm in Section \ref{algorithmevaluation}.

\ifCLASSOPTIONcaptionsoff
  \newpage
\fi



%\bibliographystyle{IEEEtran}

\bibliography{bibfile}
\bibliographystyle{unsrt}

\end{document}